%% file: main.tex
\documentclass[11pt]{article}

\usepackage[preprint]{acl}

\usepackage{times}
\usepackage{latexsym}
\usepackage{algorithm}
\usepackage{algorithmic}
\usepackage{amsmath}
\usepackage{amssymb}
\usepackage{xspace}
\usepackage{multirow}

\usepackage{newfloat}
\usepackage{listings}

\usepackage{booktabs}

\providecolor{RoyalBlue}{RGB}{65,105,225}

\newcommand{\method}{RACE\xspace}

\usepackage[T1]{fontenc}
\usepackage[utf8]{inputenc}

\usepackage{float}
\usepackage{microtype}

\usepackage{inconsolata}

\usepackage{graphicx}

\title{Reward-Adaptive Co-Evolution of Policy and Rubric}

\author{
    \textbf{Yizhu Chen\textsuperscript{1, 2}},
    \textbf{Jianguang Jiang\textsuperscript{2}},
    \textbf{Jiameng Huang\textsuperscript{1}},
    \textbf{Chuyu Fang\textsuperscript{2}},\\
    \textbf{Bojun Zhang\textsuperscript{2, 3}},
    \textbf{Zhenyu Jiao\textsuperscript{2}},
    \textbf{Xing Yuan\textsuperscript{2}},
    \textbf{Tong Lin\textsuperscript{1,\textdagger}}
    \\\\
    \textsuperscript{1}Peking University \quad \textsuperscript{2}Baidu Inc.
    \quad \textsuperscript{3}Institute of Automation, Chinese Academy of Sciences.
    \\\\
    \small{
        \textsuperscript{\textdagger}\textbf{Corresponding author:}
        \href{mailto:lintong@pku.edu.cn}{lintong@pku.edu.cn}
    }
}

\begin{document}
\maketitle

\input{sec/abs}

\input{sec/intro}

\input{sec/related}

\input{sec/preliminary}

\input{sec/obs}

\input{sec/method}

\input{sec/exp}

\input{sec/conclusions}

\input{sec/limitation}


\bibliography{custom}

\appendix

\input{sec/app}

\end{document}

%% file: sec/abs.tex
\begin{abstract}
Rubric-based rewards provide fine-grained, interpretable supervision for open-ended reinforcement learning, where holistic scalar rewards struggle to capture multidimensional response quality. Static rubrics can lose discriminative power as the policy distribution evolves, while existing policy--rubric co-evolution methods rely on globally prescribed update schedules that ignore heterogeneous reward informativeness. We introduce \method, a \textbf{R}eward-\textbf{A}daptive \textbf{C}o-\textbf{E}volution framework. At the group level, per-rollout-group training-signal routing classifies groups as informative, diversity-collapsed, or rubric-failed, updating the policy only with informative groups while caching rubric failures. At the phase level, event-triggered phase switching enters rubric optimization upon persistent reward saturation and returns when rubric-reward improvement plateaus. Rubric phases train the generator on cached pairs using a frozen judge and a scale-normalized pairwise-margin reward. Across both levels, online threshold adaptation adjusts routing and switching thresholds through proportional control. Under a unified evaluation protocol, \method achieves the highest average performance and outperforms the strongest fixed-alternation baseline on all eight tasks, while improving rubric discrimination.
\end{abstract}

% Rubric-based rewards provide fine-grained, interpretable supervision for reinforcement learning (RL) on open-ended tasks where scalar reward models fall short. Offline generated rubrics always become misaligned as the policy evolves. Self-evolving methods address this limitation by jointly updating the policy and rubric generator, yet their predefined global update schedules cannot adapt to changing reward informativeness. To address these challenges, we propose RACE, a reward-adaptive co-evolution framework that dynamically alternates policy and rubric evolution. Specifically, RACE introduces (1) \textbf{per-rollout group training signal routing}, which filters uninformative rollout groups so that only discriminative rewards drive policy updates; and (2) \textbf{event-triggered phase switching}, which dynamically determines when to alternate between policy and rubric training. Together, these mechanisms ensure the policy always trains on informative reward signal and the rubric is updated only when needed, eliminating wasted updates. Extensive experiments across reasoning and generation benchmarks demonstrate that \method outperforms existing rubric-based RL methods on at least seven of the nine benchmarks evaluated with less compute.

%% file: sec/intro.tex
\section{Introduction}
% 第一段和第二段合并
% motivate 慎用

% Reinforcement learning (RL) has substantially improved large language model
% alignment~\cite{schulman2017proximal,ouyang2022training,shao2024deepseekmath},
% but open-ended tasks remain challenging because their multidimensional quality
% cannot be adequately captured by a single scalar reward
% ~\cite{pmlr-v202-gao23h}. Rubric-based rewards instead provide fine-grained,
% interpretable supervision through explicit evaluation criteria
% ~\cite{liu2026rules,huang2025reinforcement,zhang2026chasing, shen2026rethinking}. Yet most methods rely on offline rubrics or fixed generators constructed with proprietary teachers, preference labels, or task-specific verifiers~\cite{gunjal2025rar,shen2026rethinking, viswanathan2025checklists,liu-etal-2026-openrubrics}. Because these criteria remain static during policy training, they can lose alignment and discriminative power as the output distribution evolves.

Reinforcement learning (RL) has substantially improved large language model
alignment~\cite{schulman2017proximal,ouyang2022training,shao2024deepseekmath},
yet open-ended tasks remain challenging due to the limitations of scalar rewards
in capturing multidimensional quality~\cite{pmlr-v202-gao23h}. Rubric-based
rewards address this issue with fine-grained, interpretable criteria
~\cite{liu2026rules,huang2025reinforcement,zhang2026chasing,shen2026rethinking}.
However, existing methods often rely on static rubrics or fixed generators,
which may lose alignment and discriminative power as policies evolve
~\cite{gunjal2025rar,shen2026rethinking,viswanathan2025checklists,liu-etal-2026-openrubrics}.

Rather than keeping rubrics fixed, policy--rubric co-evolution updates the policy and rubric generator in tandem, allowing each to adapt to changes induced by the other~\citep{li2026evolmselfevolvinglanguagemodels}. Existing approaches realize this idea through joint optimization within a single model~\citep{sheng2026reinforcingchainofthoughtreasoningselfevolving}, fixed-schedule alternation between policy and rubric phases~\citep{li2026evolmselfevolvinglanguagemodels}, or adversarial interaction at each training step~\citep{ding2026evorubrics}. However, these methods rely on globally prescribed update schedules and ignore prompt-level reward informativeness. Because reward degeneracy is heterogeneous, some rollout groups receive nearly identical rewards as the policy improves, causing normalized advantages to vanish (\emph{advantage collapse})~\citep{he2026advantage}. Uniform updates therefore waste policy compute on collapsed groups and rubric compute on adaptations that may be premature, delayed, or unnecessary.

To address this limitation, we introduce \method, a
\textbf{R}eward-\textbf{A}daptive \textbf{C}o-\textbf{E}volution framework that coordinates policy and rubric optimization according to reward informativeness. \method has three components. First, \textbf{per-rollout-group training-signal routing} uses reward discrimination and response diversity to route informative groups to policy optimization, cache rubric failures for rubric generator optimization, and mask diversity-collapsed groups from policy updates. Second, \textbf{event-triggered phase switching} enters rubric optimization when reward saturation persists and returns to policy optimization when rubric-reward improvement plateaus, replacing fixed alternation with state-dependent transitions. Third, the rubric generator is optimized on cached failure pairs using a frozen judge and a scale-normalized pairwise-margin reward, encouraging rubrics that restore discrimination where the current reward signal is insufficient. Online threshold adaptation adjusts the routing and phase-switching thresholds in response to evolving training dynamics. Together, these mechanisms route updates at both the rollout-group and phase levels, allowing the policy to learn only from informative rewards while adapting the rubric only where and when needed.

% Empirically, under a unified evaluation protocol spanning mathematics, code generation, general reasoning, knowledge, and instruction following, \method achieves the strongest average downstream performance among the compared methods. It also improves rubric discrimination over the strongest fixed-length alternation baseline on both JudgeBench and RewardBench2, while requiring fewer optimization steps. Analyses of rubric evolution and training dynamics further show that \method produces more specific and verifiable criteria and exhibits substantially less policy-phase advantage collapse. These results demonstrate that adapting rubric supervision to the evolving policy improves both the effectiveness and efficiency of rubric-based RL.

Empirically, under a unified evaluation protocol spanning mathematics, code generation, reasoning, knowledge, and instruction following, \method achieves the best average downstream performance among compared methods. It also improves rubric discrimination over the strongest fixed-length alternation baseline on JudgeBench and RewardBench2 with fewer optimization steps. Further analyses show that \method produces more specific, verifiable criteria and reduces policy-phase advantage collapse, demonstrating the effectiveness of adapting rubric supervision to evolving policies.

%% file: sec/related.tex
\section{Related Works}

\subsection{Rubric-Based Reinforcement Learning}

Open-ended tasks require multidimensional evaluation beyond binary correctness. Reinforcement Learning with Rubric Rewards (RLRR) decomposes response quality into explicit criteria and uses rubric-conditioned judgments as training
rewards~\citep{gunjal2025rar,chen2026holistic}. It has shown promise in science
and healthcare~\citep{gunjal2025rar}, instruction
following~\citep{huang2025reinforcement,viswanathan2025checklists,
he2025advancedifrubricbasedbenchmarkingreinforcement}, and agentic
tasks~\citep{team2025kimi}.

Most RLRR methods rely on rubrics constructed independently of policy training. Prompted generation elicits rubrics from off-the-shelf teacher models~\citep{gunjal2025rar,shen2026rethinking,jia2026openrs}, inheriting their capability and inference cost. Supervised rubric generation instead learns rubric-related components from preference labels, correctness signals, or human
annotations~\citep{sheng2026reinforcingchainofthoughtreasoningselfevolving, xu2026alternatingreinforcementlearningrubricbased,
he2025advancedifrubricbasedbenchmarkingreinforcement}, shifting the supervision bottleneck to external signals. Because these rubrics or generators are typically fixed during policy optimization, they cannot adapt to changes in the policy distribution. Motivated by the discriminative view of rubric quality~\citep{li2026evolmselfevolvinglanguagemodels}, we instead co-evolve the policy and rubric generator and allocate their updates according to observed reward informativeness.

% \footnote{RaR: gunjal2025rar; EvoLM: li2026evolmselfevolvinglanguagemodels;
% RRD: shen2026rethinking;
% RLCER: sheng2026reinforcingchainofthoughtreasoningselfevolving;
% Rubric-ARM: xu2026alternatingreinforcementlearningrubricbased
% }
\subsection{Self-Evolving LLMs}

Self-evolving LLMs and agents update their behavioral policies using their own trajectories, self-generated data, or feedback signals, enabling experience-dependent improvement~\citep{gao2026a,fang2025comprehensive, tao2024survey}.

Recent work extends self-evolution from scalar feedback to explicit natural-language rubrics. Existing methods use predefined coupling schemes: shared-parameter joint optimization in
RLCER~\citep{sheng2026reinforcingchainofthoughtreasoningselfevolving},
alternating generator--judge optimization in Rubric-ARM~\citep{xu2026alternatingreinforcementlearningrubricbased},
fixed-length policy--rubric phases in EvoLM~\citep{li2026evolmselfevolvinglanguagemodels}, and per-step dual-adapter updates in EvoRubrics~\citep{ding2026evorubrics}. Several methods also rely on external supervision or verification, including preference labels in Rubric-ARM, expert annotations in RIFL~\citep{he2025advancedifrubricbasedbenchmarkingreinforcement}, external validators in RLAC~\citep{wu2025rlac} and repeated external API calls for EvoRubrics' per-step rubric--judge interaction. Despite these differences, their update allocation is globally prescribed and does not account for rollout-group reward informativeness. Our method instead combines per-rollout-group training-signal routing with event-triggered phase switching, updating the policy only on informative groups and adjusting rubric optimization only when reward discrimination fails, without relying on external labels or task-specific verifiers.

%% file: sec/preliminary.tex
\section{Preliminary}

\paragraph{Rubric-Based Reward.}
Given an input $x$, a rubric generator $\mathcal{G}$ produces
\begin{equation}
R_x \sim \mathcal{G}(\cdot\mid x),
\qquad
R_x=\{(c_k,w_k)\}_{k=1}^{K},
\label{eq:rubric_gen}
\end{equation}
where $c_k$ and $w_k$ denote the $k$-th evaluation criterion and its weight. Let $\mathcal{J}$ denote the judge model, and let $J(x,y;R)$ denote the scalar score it assigns to response $y$ for prompt $x$ under rubric $R$. Given a policy response $y\sim\pi_\theta(\cdot\mid x)$, implicit aggregation~\citep{gunjal2025rar} yields
\begin{equation}
S_{\mathrm{policy}}(x,y)=J(x,y;R_x),
\label{eq:rubric_reward}
\end{equation}
avoiding separate judge calls for individual criteria while retaining structured supervision.

% where $J(c_k, y)$ is a scoring function (implemented by a judge $\mathcal{J}$) that evaluates to what extent the response $y$ satisfies criterion $c_k$. In practice, we adopt implicit aggregation~\citep{gunjal2025rar}: the judge receives the full rubric set $R_x$ with the response $y$ and directly produces an overall score $S(x, y) = J(x, y, R_x)$, avoiding the cost of per-criterion evaluation while preserving structured supervision.

\paragraph{Group Relative Policy Optimization.}
For each prompt $x$, the policy samples $G$ responses
$\{y_i\}_{i=1}^{G}$, yielding rewards
$r_i=S_{\mathrm{policy}}(x,y_i)$ and the reward
vector $\mathbf r=(r_1,\ldots,r_G)$. The group-normalized advantage for response $y_i$ is
\begin{equation}
    A_i = \frac{r_i - \mu(\mathbf{r})}{\sigma(\mathbf{r}) + \epsilon},
\label{eq:advantage}
\end{equation}
where $\mu(\mathbf{r})$ and $\sigma(\mathbf{r})$ denote the within-group
mean and standard deviation of the rewards, respectively, and $\epsilon>0$
ensures numerical stability. The policy is optimized by minimizing the negative clipped surrogate objective~\citep{shao2024deepseekmath}:
\begin{align}
\resizebox{0.89\linewidth}{!}{$\mathcal{L}_{\mathrm{GRPO}}(\theta)
= -\frac{1}{G}
\sum_{i=1}^{G}
\frac{1}{|y_i|}
\sum_{t=1}^{|y_i|}
\left[
\hat A_{i,t}
- \beta D_{\mathrm{KL}}^{i,t}
\left(\pi_\theta\,\|\,\pi_{\mathrm{ref}}\right)
\right],$}
\label{eq:grpo_loss}
\end{align}
where the clipped advantage term is $\hat{A}_{i,t} = \min\!\bigl(\rho_{i,t}\,A_i,\; \mathrm{clip}(\rho_{i,t},\,1{-}\epsilon_c,\,1{+}\epsilon_c)\,A_i\bigr)$, $\rho_{i,t} = \pi_\theta(y_{i,t} \mid x, y_{i,<t}) / \pi_{\theta_{\mathrm{old}}}(y_{i,t} \mid x, y_{i,<t})$ is the importance sampling ratio, and $D_{\mathrm{KL}}^{i,t}$ is the token-level KL penalty against
$\pi_{\mathrm{ref}}$, weighted by $\beta\geq0$. The rubric generator is also optimized by GRPO, with the rubric-phase reward in Eq.~\ref{eq:rubric_train_reward}.

%% file: sec/obs.tex
\section{Advantage Collapse under Fixed-Length Policy--Rubric Alternation}
\label{sec:observation}

GRPO relies on within-group reward differences. If all rollouts receive the same reward, their normalized advantages become zero (Eq.~\ref{eq:advantage}), and the prompt contributes no policy gradient---a failure mode known as \emph{advantage collapse}~\citep{he2026advantage}. We investigate whether this issue arises under fixed-length policy--rubric alternation.

\paragraph{Setup.}
Following EvoLM~\citep{li2026evolmselfevolvinglanguagemodels}, we train a shared Qwen3-8B model~\citep{yang2025qwen3} with fixed-length policy and rubric phases, both optimized with GRPO. We assess rubric characteristics and advantage collapse at steps 200, 400, 600, 800, and 1,000, and evaluate downstream performance at steps 500 and 1,000.

At each checkpoint, we evaluate 100 fixed prompts and report the fraction of rollout groups with exact advantage collapse, defined by zero reward standard deviation ($\sigma(\mathbf r)=0$). We also report JSON validity and the label-only ratio (the fraction of criteria with at most 40 characters), a proxy for short, underspecified labels. Gemini~3.5 Flash assesses semantic redundancy by identifying rubrics containing a criterion pair such that satisfying one substantially implies satisfying the other.

\paragraph{Observations.}
Table~\ref{tab:rubric_quality} reveals a mismatch between rubric quality and reward discrimination. From steps 200 to 1000, JSON validity falls from 70\% to 82\%. Also, semantic redundancy persists, and advantage collapse remains at 27--32\% through step 800 before rising to 44\% at step 1000. Thus, improved rubric quality does not necessarily yield more informative rewards as the policy evolves. Likewise, Table~\ref{tab:stagnation} shows only a marginal increase in average score from 60.8 to 61.0.

\input{tab/obs_tab}

\paragraph{Implications.}
Low reward variation may stem from either poor rubric discrimination or homogeneous policy outputs. Fixed-length alternation cannot distinguish these cases and thus cannot allocate updates accordingly.

%% file: tab/obs_tab.tex
\begin{table}[t]
\centering
\vspace{-0.5cm}
\small
\setlength{\tabcolsep}{1.2mm}
\begin{tabular}{lccccc}
\toprule
\textbf{Step}
& \textbf{Valid.(\%)}
& \textbf{Label (\%)}
& \textbf{Redund.(\%)}
& \textbf{$\sigma=0$ (\%)} \\
\midrule
% 100  & 93 & 5.2 & 11 & 21 & 31 \\
200  & 70  & 9  & 28  & 32 \\
400  & 87  & 4  & 16 & 27 \\
% 500  & 99 & 5.1 & 5  & 16 & 30 \\
600  & 99  & 5  & 23  & 27 \\
800  & 95  & 10  & 21  & 29 \\
1000 & 82  & 12  & 23 & 44 \\
\bottomrule
\end{tabular}
\caption{Rubric analysis under fixed-length alternation.}
\label{tab:rubric_quality}
% \vspace{-0.2cm}
\end{table}

\begin{table}[t]
\centering
\small
\setlength{\tabcolsep}{1mm}
\begin{tabular}{l*{6}{c}}
\toprule
Step & \textbf{GSM8K}
& \textbf{MMLU}
& \textbf{IFEval}
& \textbf{PopQA}
& \textbf{GPQA}
& \textbf{Avg.} \\
\midrule
500
& 95.8
& 85.1 & 38.6 & 30.7 & 53.6 & 60.8 \\
1000
& 95.5
& 85.6 & 38.8 & 31.2 & 54.0 & 61.0 \\
\bottomrule
\end{tabular}
\caption{Downstream performance.}
\label{tab:stagnation}
\vspace{-0.3cm}
\end{table}

%% file: sec/method.tex
\begin{figure*}[t]
\vspace{-0.5cm}
\centering
\includegraphics[width=1.0\linewidth]{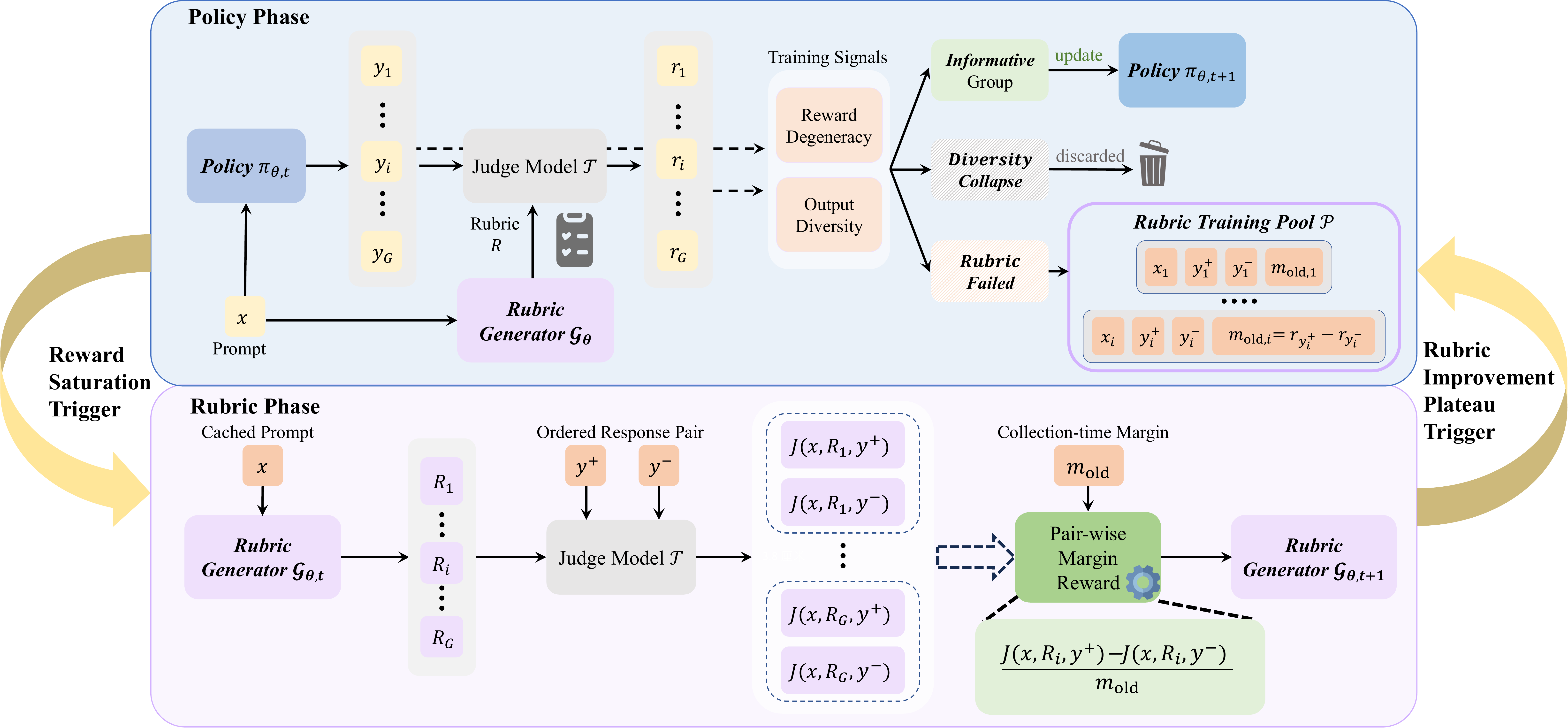}
\vspace{-0.4cm}
\caption{\textbf{Framework overview of \method.} A shared model parameterized by $\theta$ alternates between the policy $\pi_\theta$ and rubric generator $\mathcal G_\theta$. During the policy phase, $\pi_\theta$ samples $\{y_i\}_{i=1}^{G}$ for prompt $x$, and the frozen judge $\mathcal J$ scores them under the prompt-specific rubric $R$ via
Eq.~\ref{eq:rubric_reward}. \textbf{Per-rollout-group training-signal routing} uses informative groups to update the policy, masks diversity-collapsed groups, and caches rubric-failed entries $(x,y^+,y^-,m_{\mathrm{old}})$ in $\mathcal P$. During the rubric phase, $\mathcal G_\theta$ samples candidate rubrics $\{R_i\}_{i=1}^{G}$ for each cached entry, each evaluated on the same pair $(y^+,y^-)$ using the \textbf{scale-normalized pairwise-margin reward} in Eq.~\ref{eq:rubric_train_reward}. \textbf{Event-triggered phase switching} moves to rubric optimization upon persistent reward saturation and back upon a rubric-reward plateau.}
\label{fig:framework}
\vspace{-0.3cm}
\end{figure*}

\section{Method}
\label{sec:method}

%% ============================================================

% Figure~\ref{fig:framework} provides an overview of \method. A single model with shared parameters $\theta$ alternates between the policy $\pi_\theta$ and the rubric generator $\mathcal G_\theta$. The framework has three components. First, per-rollout-group training-signal routing suppresses policy updates from reward-degenerate groups and routes rubric-failed cases to a rubric training pool $\mathcal P$ for targeted rubric adaptation. Second, event-triggered switching coordinates policy and rubric optimization according to the observed training state. Third, a scale-normalized multi-pair reward trains the rubric generator on bundles of cached response pairs. Both roles are optimized with GRPO using phase-specific rewards. Algorithm~\ref{alg:tempo} summarizes the procedure.\footnote{need to change}

Figure~\ref{fig:framework} illustrates the overall framework of \method. A single model with shared parameters $\theta$ serves alternately as the policy $\pi_\theta$ and the rubric generator $\mathcal G_\theta$, with both roles optimized by GRPO under phase-specific rewards. The framework operates at three levels. At the rollout-group level, training-signal routing retains informative groups for policy optimization and stores rubric-failed cases in a FIFO pool $\mathcal P$. At the phase level, event-triggered switching enters rubric optimization when reward saturation persists and returns to policy optimization when rubric improvement plateaus; during the rubric phase, candidate rubrics are trained on cached response pairs using a scale-normalized pairwise margin reward. Finally, projected proportional controllers adapt the routing and phase-switching thresholds as training evolves. Algorithm~\ref{alg:tempo} in App.~\ref{app:algorithm} summarizes the full procedure.

% motivation

\subsection{Per-Rollout-Group Training-Signal Routing}
\label{sec:routing}

Rollout groups vary in reward informativeness, and low reward variation may arise from either homogeneous policy responses or insufficient rubric discrimination. To distinguish these,  \method assigns each group to one of three modes using its $G$ responses and reward vector $\mathbf r=\{r_i\}_{i=1}^{G}$. Informative groups update the policy;
diversity-collapsed groups are masked from the policy loss; and rubric-failed groups are masked and cached in a FIFO pool $\mathcal P$ as targeted training data for subsequent rubric generator optimization.

% Routing uses the reward-variance threshold $\sigma_{\mathrm{rt}}^\star$, the tie-rate threshold $\tau_{\mathrm{tie}}^\star$, and the diversity thresholds $\operatorname{cv}^\star$ and $d^\star$. Their initial values are calibrated using early training statistics, as detailed in Appendix 1. TRQC adapts the two reward-side thresholds online to maintain a target quarantine rate, while the diversity thresholds remain fixed.

\input{tab/mode}

\paragraph{Routing Statistics.}
For each rollout group, we measure reward discrimination using the reward
standard deviation $\sigma(\mathbf r)$ and tie rate
\begin{equation}
\tau_{\mathrm{tie}}(\mathbf r)
=
\frac{
\left|\{(i,j):i<j,\ |r_i-r_j|<\epsilon_{\mathrm{tie}}\}\right|
}{
G(G-1)/2
},
\label{eq:tie_rate}
\end{equation}
where $\epsilon_{\mathrm{tie}}>0$ defines indistinguishable rewards. We measure surface-form diversity using the distinct-trigram ratio:
\begin{equation}
d_{\mathrm{tri}}
=
\frac{|\mathcal T_{\mathrm{unique}}|}
     {|\mathcal T_{\mathrm{all}}|},
\label{eq:surface_diversity}
\end{equation}
where $\mathcal T_{\mathrm{all}}$ is the multiset of all trigrams across the $G$ responses, and $\mathcal T_{\mathrm{unique}}$ is its set of unique trigrams. To approximate semantic diversity, we encode each response using a frozen encoder, with last-token pooling and $\ell_2$ normalization:
\begin{align}
\resizebox{0.87\linewidth}{!}{$\mathbf z_i
=
\frac{\mathcal E(y_i)}{\|\mathcal E(y_i)\|_2},
\
d_{\mathrm{emb}}
=
\frac{2}{G(G-1)}
\sum_{i<j}
\left(1-\mathbf z_i^\top\mathbf z_j\right).$}
\label{eq:embedding_diversity}
\end{align}
Here, $d_{\mathrm{emb}}$ is the mean pairwise cosine distance within the rollout group.

We apply the routing rules in Table~\ref{tab:modes} sequentially. A group is first classified as reward-degenerate when
\begin{equation}
\sigma(\mathbf r)\leq\sigma_{\mathrm{rt}}^\star
\quad\text{or}\quad
\tau_{\mathrm{tie}}(\mathbf r)\geq\tau_{\mathrm{tie}}^\star.
\label{eq:reward_degenerate}
\end{equation}
A reward-degenerate group is \emph{diversity-collapsed} if both surface and semantic diversity are low:
\begin{equation}
d_{\mathrm{tri}}\leq d_{\mathrm{tri}}^{\star}
\;\wedge\;
d_{\mathrm{emb}}\leq d_{\mathrm{emb}}^{\star}.
\label{eq:surface_form_collapse}
\end{equation}
Diversity-collapsed groups are filtered from policy optimization, while remaining reward-degenerate groups are labeled \emph{rubric-failed}, filtered, and added to rubric training pool $\mathcal P$ for rubric-generator optimization, indicating weak rubric discrimination over varied responses. Non-degenerate groups are \emph{informative} and proceed with standard GRPO.

% Diversity-collapsed groups are masked from the policy loss. Remaining reward-degenerate groups are classified as \emph{rubric-failed}, masked, and added to the rubric training pool $\mathcal P$ for rubric-generator optimization. This label indicates weak rubric discrimination over varied responses. Non-degenerate groups are \emph{informative} and proceed with standard GRPO.

All starred thresholds are initialized from statistics collected during the first 50 policy-training steps. The reward-side thresholds $\sigma_{\mathrm{rt}}^\star$ and $\tau_{\mathrm{tie}}^\star$ are subsequently adapted online.

\paragraph{Rubric Training Pool.}
%The FIFO pool $\mathcal P$ stores rubric-failed groups. 
The FIFO pool $\mathcal P$ stores entries constructed from
rubric-failed groups. For a group with responses
$\{y_i\}_{i=1}^{G}$, let
$r_i=J(x,y_i;R_{\mathrm{old}})$ denote the reward assigned
using the collection-time rubric $R_{\mathrm{old}}$. We select
\begin{equation}
i^+ = \operatorname*{arg\,max}_{i\in\{1,\ldots,G\}} r_i,
\
i^- = \operatorname*{arg\,min}_{i\in\{1,\ldots,G\}\setminus\{i^+\}} r_i,
\label{eq:pair_selection}
\end{equation}
using a deterministic rule to break ties, and set
$y^+=y_{i^+}$ and $y^-=y_{i^-}$. Each pool entry contains a prompt $x$, an ordered response pair $(y^+,y^-)$, and its collection-time margin
\begin{equation}
m_{\mathrm{old}}
=
J(x,y^+;R_{\mathrm{old}})
-
J(x,y^-;R_{\mathrm{old}}),
\label{eq:old_margin}
\end{equation}
%where $R_{\mathrm{old}}$ is the rubric used at collection. 
The cached margin supports Eq.~\ref{eq:rubric_train_reward} without re-evaluating the old rubric,
while $\mathcal P$ focuses rubric optimization on prompts with weak collection-time reward
discrimination.

%  During the subsequent rubric phase, these prompts serve as targeted training data, since they are exactly the cases where the current rubric cannot discriminate.

% \footnote{\huang{This paper contains a large number of symbols, so I suggest assigning natural-language names to all symbols that will be referenced later.}}

% \footnote{all changes to policy model}

% However, a fixed alternation period is suboptimal: when the rubric loses discriminative power early, the policy trains on uninformative signal; when it remains effective, switching wastes rubric-phase compute. \method replaces fixed-period alternation with event-triggered transitions that fire only when the reward signal demands it.

\subsection{Event-Triggered Phase Switching}
\label{sec:switching}

As shown in Section~\ref{sec:observation}, fixed-length alternation may continue policy optimization after rewards lose discrimination. \method enters the rubric phase when reward saturation persists across rollout groups and returns when rubric improvement plateaus.

\paragraph{Policy $\to$ Rubric: Reward-Saturation Trigger.}
At policy step $t$, we measure the fraction of saturated rollout groups:
\begin{equation}
\resizebox{0.87\linewidth}{!}{$f_{\mathrm{sat}}^{t}
=
\frac{1}{N_{\mathrm{batch}}}
\sum_{j=1}^{N_{\mathrm{batch}}}
\mathbb{1}\!\left[
\sigma(\mathbf r_j)\leq\sigma_{\mathrm{sw}}^\star
\;\wedge\;
\mu(\mathbf r_j)\geq\mu_{\mathrm{sat}}
\right],$}
\label{eq:reward_saturation_rate}
\end{equation}
where $N_{\mathrm{batch}}$ is the number of groups, $\mathbf r_j=(r_{j,1},\ldots,r_{j,G})$ denotes the $G$ rewards of group $j$, and $\sigma(\cdot)$ and $\mu(\cdot)$ represent reward variation and mean. The thresholds $\sigma_{\mathrm{sw}}^\star$ and $\mu_{\mathrm{sat}}$ specify low variation and high reward, respectively.
% where $N_{\mathrm{batch}}$ is the number of groups in the batch, $\mathbf r_j=(r_{j,1},\ldots,r_{j,G})$ contains the $G$ rewards of group $j$, and $\sigma(\cdot)$ and $\mu(\cdot)$ denote standard deviation and mean. The thresholds $\sigma_{\mathrm{sw}}^\star$ and $\mu_{\mathrm{sat}}$ define low reward variation and high mean reward, respectively.

To smooth transient batch fluctuations, we maintain an exponential moving average (EMA):
\begin{equation}
\hat f_{\mathrm{sat}}^{t}
=
\alpha f_{\mathrm{sat}}^{t}
+
(1-\alpha)\hat f_{\mathrm{sat}}^{t-1},
\label{eq:saturation_ema}
\end{equation}
where $\alpha\in(0,1)$ is the EMA smoothing factor. The controller enters the
rubric phase when
$\hat f_{\mathrm{sat}}^{t}\geq\tau_{\mathrm{sat}}^{\mathrm P}$, where
$\tau_{\mathrm{sat}}^{\mathrm P}$ is the phase-level saturation-rate threshold.
Thus, $\sigma_{\mathrm{sw}}^\star$ and $\mu_{\mathrm{sat}}$ identify saturation
within each group, whereas $\tau_{\mathrm{sat}}^{\mathrm P}$ requires it to
persist across batches before switching.

\paragraph{Rubric $\to$ Policy: Rubric-Improvement Plateau Trigger.}
During the rubric phase, we monitor whether optimization continues to yield meaningful reward gains. Let $\bar r_t$ denote the mean rubric reward at step $t$. We measure the improvement between two consecutive windows of size $W$:
\begin{equation}
\Delta_W^{t}
=
\frac{1}{W}\sum_{u=t-W+1}^{t}\bar r_u
-
\frac{1}{W}\sum_{u=t-2W+1}^{t-W}\bar r_u.
\label{eq:rubric_plateau}
\end{equation}
Once both windows are available, the controller returns to the policy phase when $\Delta_W^{t} \leq \tau_{\mathrm{imp}}^{\mathrm{R}}$, where $\tau_{\mathrm{imp}}^{\mathrm{R}}$ is the minimum reward improvement required to continue the rubric phase.

% \input{tab/algorithm}
% Conversely, when the rubric generator $$\mathcal{G}_\theta$$ stops improving, continued rubric training wastes compute. We detect this by checking whether the step-level reward $r_t$ is still increasing over a sliding window of size $W$:
% \begin{equation}
%     \begin{cases}
%     \text{If} \ \frac{1}{W}\sum_{i=t-W+1}^{t} r_i - \frac{1}{W}\sum_{i=t-2W+1}^{t-W} r_i < \epsilon_{\text{abs}},&\text{switch;} \\
%     \text{Otherwise},&\text{continue.}
%     \end{cases}
% \end{equation}
% When this condition holds, the rubric has converged and training switches back to the policy phase. All threshold values are reported in Appendix 1.

\paragraph{Rubric Generator Optimization.}
Given an entry $(x,y^+,y^-,m_{\mathrm{old}})$ sampled from rubric training pool $\mathcal P$, the rubric generator $\mathcal{G}_\theta$  samples a GRPO group of $G$ candidate rubrics $\{R_i\}_{i=1}^{G}$, where $R_i\sim\mathcal G_\theta(\cdot\mid x)$. All candidates are evaluated on the same ordered response pair. For each $R_i$, the frozen judge computes
\begin{equation}
m_i
=
J(x,y^+;R_i)
-
J(x,y^-;R_i).
\label{eq:new_margin}
\end{equation}
The candidate receives the scale-normalized pairwise reward
\begin{align}
\resizebox{0.87\linewidth}{!}{$S_{\mathrm{rubric}}(x,R_i)
=
\operatorname{clip}\!\left(
\frac{m_i}
{\max(|m_{\mathrm{old}}|,\epsilon_{\mathrm{norm}})},
-c,c
\right).$}
\label{eq:rubric_train_reward}
\end{align}
where $m_{\mathrm{old}}$ is the cached historical margin defined in Eq.~\ref{eq:old_margin}. The historical margin normalizes pair-specific reward scales, while a negative
$m_i$ penalizes reversal of the cached ordering. We set
$\epsilon_{\mathrm{norm}}=0.05$ and $c=1.0$ to stabilize normalization and bound the reward. The reward measures separation along the cached direction relative to its historical scale.

\paragraph{Phase-Specific GRPO Rewards.}
Both roles are optimized with the clipped GRPO objective in
Eq.~\ref{eq:grpo_loss}; only the reward function differs between phases:
\begin{equation}
\resizebox{0.87\linewidth}{!}{$r_i
=
\begin{cases}
S_{\mathrm{policy}}(x,y_i),
& \text{during the policy phase}, \\[2pt]
S_{\mathrm{rubric}}(x,R_i),
& \text{during the rubric phase}.
\end{cases}$}
\label{eq:phase_rewards}
\end{equation}
Here, $i\in\{1, \ldots, G\}$ indexes responses in the policy phase and candidate rubrics in the rubric phase. The corresponding rewards are defined in Eqs.~\ref{eq:rubric_reward} and~\ref{eq:rubric_train_reward},
respectively.

\subsection{Online Threshold Adaptation}
\label{sec:threshold_adaptation} 

% As the policy and reward distributions evolve, fixed thresholds can become miscalibrated. We therefore adapt them online using two proportional controllers. The Target-Rate Quarantine Controller (TRQC) adapts $\sigma_{\mathrm{rt}}^\star$ and $\tau_{\mathrm{tie}}^\star$ to regulate per-rollout-group training-signal routing, while the Adaptive Phase-Switching Controller adjusts $\tau_{\mathrm{sat}}^{\mathrm P}$ and $\tau_{\mathrm{imp}}^{\mathrm R}$ to regulate transitions between policy and rubric optimization.

As the policy and reward distributions evolve, fixed thresholds can become miscalibrated. We therefore adapt them online using two proportional controllers. The Target-Rate Quarantine Controller (TRQC) adapts $\sigma_{\mathrm{rt}}^\star$ and $\tau_{\mathrm{tie}}^\star$ for rollout-group training-signal routing, while the Adaptive Phase-Switching Controller adjusts $\tau_{\mathrm{sat}}^{\mathrm P}$ and $\tau_{\mathrm{imp}}^{\mathrm R}$ for policy--rubric phase transitions.

\paragraph{Target-Rate Quarantine Controller.}
At policy step $t$, we measure the quarantine rate
\begin{equation}
q^{t}
=
n_{\mathrm{quarantined}}^{t}
\,/\,
n_{\mathrm{processed}}^{t},
\label{eq:trqc_rate}
\end{equation}
where quarantine covers diversity-collapsed and rubric-failed groups, with
target quarantine rate $q^*$.

We maintain the smoothed tracking error
\begin{equation}
\resizebox{0.87\linewidth}{!}{$\hat q^{t}
=
(1-\alpha_q)\hat q^{(t-1)}+\alpha_q q^{t},
\
e_q^{t}
=
\frac{\hat q^{t}-q^*}{\max(q^*,\epsilon_q)},$}
\label{eq:trqc_ema}
\end{equation}
where $\hat q^{(0)}=0$, $\alpha_q\in(0,1)$ is the smoothing factor, and
$\epsilon_q>0$ ensures numerical stability. The routing thresholds are updated
as
\begin{equation}
\begin{aligned}
\sigma_{\mathrm{rt}}^{(t+1)}
&=
\operatorname{clip}\!\left(
\sigma_{\mathrm{rt}}^{t}-\eta_\sigma e_q^{t},\,0,\,1
\right),\\
\tau_{\mathrm{tie}}^{(t+1)}
&=
\operatorname{clip}\!\left(
\tau_{\mathrm{tie}}^{t}+\eta_\tau e_q^{t},\,0,\,1
\right),
\end{aligned}
\label{eq:trqc_update}
\end{equation}
where $\eta_\sigma,\eta_\tau>0$ are fixed controller gains. When
$\hat q^{t}>q^*$, the update makes quarantine more selective; it reverses direction when $\hat q^{t}<q^*$.

\paragraph{Adaptive Phase-Switching Controller.}
We adapt the switching thresholds toward target policy- and rubric-phase
lengths $L^*$ and $M^*$. Let $\bar L^{(k)}$ and $\bar M^{(k)}$ denote the
mean lengths of the most recent $W_{\mathrm P}$ saturation-terminated policy
phases and $W_{\mathrm R}$ plateau-terminated rubric phases, respectively.
Their normalized duration errors are
\begin{equation}
e_{\mathrm P}^{(k)}
=
\frac{\bar L^{(k)}-L^*}{L^*},
\quad
e_{\mathrm R}^{(k)}
=
\frac{\bar M^{(k)}-M^*}{M^*}.
\label{eq:phase_length_errors}
\end{equation}
Once the corresponding window is available, we update
\begin{equation}
\begin{aligned}
\tau_{\mathrm{sat}}^{\mathrm P,(k+1)}
&=
\Pi_{\mathcal I_{\mathrm P}}\!\left(
\tau_{\mathrm{sat}}^{\mathrm P,(k)}
-\eta_\tau^{\mathrm P}e_{\mathrm P}^{(k)}
\right),\\
\tau_{\mathrm{imp}}^{\mathrm R,(k+1)}
&=
\Pi_{\mathcal I_{\mathrm R}}\!\left(
\tau_{\mathrm{imp}}^{\mathrm R,(k)}
+\eta_\tau^{\mathrm R}e_{\mathrm R}^{(k)}
\right),
\end{aligned}
\label{eq:adaptive_phase_thresholds}
\end{equation}
where $\Pi_{\mathcal I}$ projects a threshold onto its valid range. The opposite signs follow from the trigger definitions: each update makes switching more likely when the corresponding phase exceeds its target duration. The targets $L^*$ and $M^*$ serve as soft operating points rather than fixed phase lengths; transitions remain governed by reward saturation and rubric-reward improvement.

\input{tab/mainexp}

%% file: tab/mode.tex
% \begin{table}[t]
% \centering
% \small
% \setlength{\tabcolsep}{0.5mm}
% \begin{tabular}{lll}
% \toprule
% \textbf{Mode} & \textbf{Diversity Gate} & \textbf{Action} \\
% \midrule
% \multicolumn{3}{l}{
% \textit{Reward-degenerate:}
% $\sigma(\mathbf{r}) \leq \sigma_{\mathrm{rt}}^\star$
% or $\tau_{\mathrm{tie}}(\mathbf{r}) \geq \tau_{\mathrm{tie}}^\star$
% } \\
% \ Surface-collapsed
% & low ($\operatorname{cv}_{\ell} \leq \operatorname{cv}^\star$, $d_{\mathrm{tri}} \leq d^\star$)
% & zero grad \\
% \ Rubric-failed
% & high (otherwise)
% & \shortstack[l]{zero grad; add to\\rubric training pool $\mathcal{P}$} \\
% \midrule
% Informative
% & ---
% &  Update \\
% \bottomrule
% \end{tabular}
% \caption{Per-rollout-group routing rules. The surface-form diversity gate heuristically attributes reward degeneracy to policy-side surface-form collapse (surface-collapsed) or rubric failure; it does not detect semantic collapse.}
% \label{tab:modes}
% \end{table}

\begin{table}[t]
\centering
\vspace{-0.2cm}
\small
\setlength{\tabcolsep}{0.4mm}
\resizebox{\linewidth}{!}{\begin{tabular}{lll}
\toprule
\textbf{Mode} & \textbf{Condition} & \textbf{Action} \\
\midrule
Diversity-collapsed
& Low surface/semantic diversity
& Mask \\

Rubric-failed
& Degenerate; otherwise
& Mask; add to $\mathcal P$ \\

Informative
& Discriminative
& standard GRPO \\
\bottomrule
\end{tabular}}
\caption{Per-rollout-group training-signal routing.}
\label{tab:modes}
\vspace{-0.4cm}
\end{table}

%% file: tab/mainexp.tex
\begin{table*}[t]
\vspace{-0.5cm}
\centering
\small
\resizebox{\linewidth}{!}{
\setlength{\tabcolsep}{0.8mm}
\begin{tabular}{lccccccccc}
\toprule
\textbf{Method}
& \textbf{GSM8K}
& \textbf{MBPP+}
& \textbf{BBH}
& \textbf{MMLU}
& \textbf{PopQA}
& \textbf{GPQA}
& \textbf{Zebra}
& \textbf{AGI-E}
& \textbf{Avg} \\
\midrule

\multicolumn{10}{l}{\textit{Prompted Rubrics Source}} \\
\quad GPT-4.1
& 95.7 & 65.1 & \underline{71.7}
& 84.8 & 30.7 & 54.9
& 81.9 & 82.9 & 71.0 \\

\quad Qwen3-8B
& 95.5 & 58.4 & 66.8
& 83.4 & 31.7 & \textbf{56.9}
& \textbf{84.3} & 83.9 & 70.1 \\

\midrule
\multicolumn{10}{l}{\textit{Scalar Reward Model}} \\
\quad Scalar RM (Skywork-RM-V2)
& 95.6 & 66.2 & 54.9
& 82.0 & 25.2 & 46.2
& 56.4 & 81.6 & 63.5 \\

\midrule
\multicolumn{10}{l}{\textit{Offline Rubric Generation}} \\
\quad RaR
& 95.7 & \underline{68.6} & 62.5
& 84.8 & 32.2 & 52.5
& \underline{84.1} & 87.3 & 71.0 \\

\quad RRD
& 95.3 & 59.2 & 69.4
& 84.8 & 32.6 & 56.0
& 82.6 & \textbf{88.9} & 71.1 \\

\midrule
\multicolumn{10}{l}{\textit{Co-evolution Methods}} \\
\quad RLCER
& 94.0 & 67.9 & 68.6
& 84.8 & 30.4 & 49.6
& 69.1 & 84.3 & 68.6 \\

\quad RUBRIC-ARM
& 95.6 & 62.3 & 64.4
& 83.8 & 32.8 & \textbf{56.9}
& 83.1 & 87.2 & 70.8 \\

\quad EvoLM
& \underline{95.8} & 68.5 & 67.6
& \underline{85.3} & 30.5 & 54.2
& 82.3 & 86.4 & \underline{71.3} \\

\quad \textbf{\method (ours)}
& \textbf{95.9} & \textbf{68.7} & \textbf{72.3}
& \textbf{86.1} & \textbf{33.8} & \underline{56.1}
& 83.9 & \underline{87.5} & $\textbf{73.0}$ \\

% \midrule
% \multicolumn{10}{l}{\textit{Judge-capacity analysis}} \\
% \quad EvoLM (Qwen3-8B Judge)
% & 95.7 & 62.1 & 68.8
% & 82.4 & 32.9 & 56.7
% & 83.6 & 87.6 & 71.2 \\

% \quad \method (Qwen3-8B Judge)
% & 96.1 & 62.5 & 69.7
% & 82.6 & 33.4 & 57.6
% & 83.8 & 88.2 & 71.7 \\

\bottomrule
\end{tabular}}
\caption{Downstream policy evaluation. Avg. is the unweighted mean; \textbf{Bold} and \underline{underlined} values indicate the best and second-best results, respectively.}

\label{tab:main_results}
\vspace{-0.4cm}
\end{table*}

%% file: sec/exp.tex
\section{Experiments}

\input{tab/rewardbench}

\subsection{Setup}

\paragraph{Data.}
We draw training prompts from the Tulu~3 preference
mixture~\citep{lambert2024tulu}. After deduplication, the resulting 271K prompts cover dialogue, instruction following, mathematics, code, scientific literature, and persona-based synthesis. We use only the prompts, without reference responses or externally generated rubrics.

% \paragraph{Implementation Details.}
% Qwen3-8B~\citep{yang2025qwen3} parameterizes both the
% policy $\pi_\theta$ and rubric generator $\mathcal{G}_\theta$. The two roles share parameters and are optimized in alternating phases. Following~\citet{li2026evolmselfevolvinglanguagemodels}, a frozen Qwen3-1.7B judge $\mathcal{J}$ provides rubric-based scores throughout training. Both roles are optimized with GRPO~\citep{shao2024deepseekmath}, using a learning rate of $10^{-6}$ and a KL coefficient of $0.001$. Each policy batch contains $N_{\mathrm{batch}}=64$ unique prompts with $G=8$ responses per prompt. During the rubric training phase, each GRPO group contains $G$ candidate rubrics evaluated on the same cached comparison set.

\paragraph{Implementation Details.}
Qwen3-8B~\citep{yang2025qwen3} parameterizes both the
policy $\pi_\theta$ and rubric generator $\mathcal{G}_\theta$. The two roles share parameters and are optimized in alternating phases. Following~\citet{li2026evolmselfevolvinglanguagemodels}, a frozen Qwen3-1.7B judge $\mathcal{J}$ provides rubric-based scores throughout training. Both roles are optimized with GRPO~\citep{shao2024deepseekmath}, using a learning rate of $10^{-6}$ and a KL coefficient of $0.001$. Each policy batch contains $N_{\mathrm{batch}}=64$ unique prompts with $G=8$ responses per prompt. During rubric training, each GRPO group contains $G$ candidate rubrics evaluated on same cached comparison set.

\paragraph{Downstream Policy Evaluation.}
Following the OLMo3-Adapt suite~\citep{olmo2025olmo}, we evaluate mathematical
reasoning on GSM8K~\citep{cobbe2021training}; code generation on MBPP+~\citep{liu2023evalplus}; general reasoning on
BBH~\citep{suzgun-etal-2023-challenging}, GPQA~\citep{rein2023gpqa},
ZebraLogic~\citep{lin2025zebralogic}, and
AGI-Eval~\citep{zhong-etal-2024-agieval} and knowledge on
MMLU~\citep{hendrycks2020mmlu} and
PopQA~\citep{mallen-etal-2023-trust}. For controlled comparison, all methods share the policy initialization, training data, frozen Qwen3-1.7B judge, and a budget of 500 policy updates.
% Qwen3-8B judge results are provided in App.~\ref{}.

\subsection{Main Comparisons}

\input{tab/ablation}
We compare \method with four baseline categories:
(i)~\emph{prompted rubrics}, including \textit{GPT-4.1 prompted} (a strong proprietary-teacher baseline) and \textit{Qwen3-8B prompted} (matching our rubric-generator backbone), both of which generate static rubrics via prompting to guide policy optimization; (ii)~\emph{scalar reward modeling}: Skywork-RM-V2; (iii)~\emph{offline rubric generation}: RaR~\citep{gunjal2025rar}, which generates rubrics with GPT-4.1, and RRD~\citep{shen2026rethinking}, which recursively refines and reweights rubric criteria; and (iv)~\emph{policy--rubric co-evolution}: RLCER~\citep{sheng2026reinforcingchainofthoughtreasoningselfevolving}, which jointly optimizes shared-parameter reasoner and rubricator roles, Rubric-ARM~\citep{xu2026alternatingreinforcementlearningrubricbased}, which alternates rubric-generator and judge updates using preference labels, and EvoLM~\citep{li2026evolmselfevolvinglanguagemodels}, which alternates between fixed 50-step policy and rubric phases.

Table~\ref{tab:main_results} reports downstream policy performance on eight benchmarks. \method achieves the highest unweighted arithmetic mean of 73.0 under this main-table protocol. It obtains the best results on GSM8K, MBPP+, BBH, MMLU, and PopQA, and ranks second on AGI-E. Compared with the strongest fixed-alternation baseline, EvoLM, \method improves on all eight main-table benchmarks, with the largest gains on BBH ($+4.7$), PopQA ($+3.3$), and GPQA ($+1.9$). \method also outperforms GPT-4.1- and Qwen3-8B-prompted rubrics by 2.0 and 2.9 points on average, respectively, and surpasses the scalar reward model by 9.5 points. These gains demonstrate the effectiveness of reward-adaptive policy--rubric co-evolution.

IFEval is reported separately in App.~\ref{app:ifeval}. We avoid direct comparison with EvoLM's reported IFEval score because the default \texttt{olmo3:adapt} protocol evaluates the full Qwen3 completion, including the \texttt{<think>...</think>} preamble, where internal reasoning may trigger violations unrelated to the final response. When scoring only the final response, \method achieves 78.2 versus 77.6 for reproduced EvoLM.

% Beyond policy performance, Table~\ref{tab:judge_rewardbench} evaluates rubric discrimination on RewardBench2~\citep{malik2025rewardbench} and JudgeBench~\citep{tan2025judgebench} under the same Qwen3-1.7B judge. JudgeBench Avg. denotes the unweighted mean over four domains (Knowledge, Reasoning, Math, and Coding), excluding the Overall score, yielding 32.2 for EvoLM and 35.0 for \method. RewardBench2 shows similar improvements, with six-dimension averages of 46.0 and 49.2. Thus, \method achieves stronger discrimination under controlled judge capacity. Comparisons with other methods are provided in App.~\ref{app:full_rewardbench}.

Beyond policy performance, Table~\ref{tab:judge_rewardbench} evaluates rubric discrimination on RewardBench2~\citep{malik2025rewardbench} and JudgeBench~\citep{tan2025judgebench} under the frozen Qwen3-1.7B judge. JudgeBench Avg. is the unweighted mean of its four domains (Knowledge, Reasoning, Math, and Coding), excluding the separately reported Overall score: 32.2 for EvoLM and 35.0 for \method. On RewardBench2, the corresponding six-dimension averages are 46.0 and 49.2. The final column additionally reports the corresponding averages under GPT-5.4 judging, where \method scores 62.2 versus 57.9 for EvoLM on JudgeBench and 65.7 versus 63.4 on RewardBench2. Thus, \method shows stronger rubric discrimination under both judge settings. Comparisons with other methods are provided in App.~\ref{app:full_rewardbench}.

\paragraph{Rubric Evolution Over Training.}
To examine rubric evolution under \method, we analyze criteria across math, emotional support, scientific explanation, and constrained writing. On 100 fixed prompts, we measure the proportions of label-only and specific criteria (containing concrete expected values), along with mean criterion length.

% We find that \method substantially reduces label-only criteria from 20.9\% to 9.9\%, while increasing specificity from 10.1\% to 18.0\% and average criterion length from 56.9 to 62.6 characters. At convergence, it also surpasses EvoLM in label-only rate (9.9\% vs.\ 11.8\%) and specificity (18.0\% vs.\ 14.5\%). These results suggest that \method enriches the content of individual criteria, making them more concrete and better suited for generating discriminative reward signals. The fixed-prompt examples, aggregate statistics, and metric definitions are provided in App.~\ref{app:rubric_examples}.

We find that \method substantially reduces label-only criteria from 20.9\% to 9.9\%, while increasing specificity from 10.1\% to 18.0\% and average criterion length from 56.9 to 62.6 characters. At convergence, it also surpasses EvoLM in label-only rate (9.9\% vs.\ 11.8\%) and specificity (18.0\% vs.\ 14.5\%). These changes are consistent with rubrics becoming more concrete and potentially more useful for producing discriminative reward signals. Direct evidence from RewardBench2 and JudgeBench is reported above. Details and analyses of rubric evolution and reward behavior are provided in App.~\ref{app:rubric_examples}.

Notably, advantage collapse persists under \method but shows no sustained increase during training. In later stages of training, the collapse rate remains between 22.4\% and 24.6\%, with an overall rate of 24.5\%. For context, the fixed-length alternation analysis of EvoLM in Sec.~\ref{sec:observation} reports rates of 27--32\% through step 800, rising to 44\% at step 1{,}000. Further analysis in App.~\ref{app:delta_analysis} shows that advantage collapse is predominantly associated with high-score saturation.

\subsection{Ablation Study}
\label{sec:ablations}

Table~\ref{tab:ablations} ablates three components of \method:
per-rollout-group training-signal routing (\emph{routing}), event-triggered phase switching (\emph{switching}), and online threshold adaptation (\emph{adaptation}), reporting RewardBench2 (RB2) and average policy performance over eight downstream benchmarks.

% \paragraph{Per-Rollout-Group Routing.}
% We separate routing's roles in masking uninformative policy gradients and collecting targeted rubric-training data. \textit{No Gradient Mask} retains failure-targeted rubric sampling pool but allows all groups to update the policy, whereas \textit{Uniform Rubric Sampling} retains masking but replaces the rubric training pool with uniformly sampled data. Removing masking causes the larger policy drop (73.0 to 70.8), while uniform sampling more strongly degrades RB2 (49.2 to 47.9). The \textit{Filter-only} variant retains alternating policy and rubric optimization but uses an independent reward-variance filter and a broader rubric training pool, reducing Policy Avg. to 70.9. These results show that both gradient masking and targeted rubric training contribute to the full method's gains.

\paragraph{Per-Rollout-Group Routing.}
As described in Section~\ref{sec:routing}, routing filters uninformative groups from policy updates and stores rubric-failed groups in the targeted rubric pool $\mathcal P$. \textit{No Gradient Filtering} retains $\mathcal P$ but updates the policy with all groups, whereas \textit{Uniform Rubric Sampling} retains filtering but replaces $\mathcal P$ with uniformly sampled data. Removing filtering lowers Policy Avg. from 73.0 to 70.8, while uniform sampling lowers RB2 from 49.2 to 47.9. The \textit{Filter-only} variant retains alternating policy and rubric optimization but replaces adaptive routing with an advantage-collapse filter alone, reducing Policy Avg. to 70.9. These results show that both group filtering and targeted rubric training contribute to the full method's gains.

\paragraph{Event-Triggered Phase Switching.}
Replacing reward-dependent transitions with random or fixed alternation lowers policy performance to 71.4 and 71.3 and RB2 to 45.3 and 44.3, respectively. These are the largest RB2 degradations, showing that routing alone is insufficient and that aligning phase transitions with reward state is critical for maintaining discriminative rubrics.

% \paragraph{Event-Triggered Phase Switching.}
% Random or fixed alternation reduces policy scores to 71.4/71.3 and RB2 to 45.3/44.3, respectively. These largest RB2 drops indicate that routing alone is insufficient, highlighting the importance of reward-dependent phase transitions.

% \paragraph{Online Threshold Adaptation.}
% \textit{Static Thresholds} fixes both routing and switching thresholds at their initial values. This variant remains competitive on several individual benchmarks but reduces the average policy score to 67.9 and RB2 to 47.1. Thus, online calibration provides a modest aggregate improvement, complementing the larger gains from routing and event-triggered switching.

\paragraph{Online Threshold Adaptation.}
\textit{Static Thresholds} fixes both routing and switching thresholds at their initial values. This variant remains competitive on several individual benchmarks but reduces the average policy score to 72.1 and RB2 to 47.1. Thus, online calibration provides a modest aggregate improvement, complementing the larger gains from routing and event-triggered switching. Target sensitivity is reported in App.~\ref{sec:hyperparameter_sensitivity}.

% \paragraph{Online Threshold Adaptation.}
% \textit{Static Thresholds} fixes routing and switching thresholds at their initial values, reducing Policy Avg.\ to 72.1 and RB2 to 47.1. Online adaptation thus yields modest gains beyond routing and event-triggered switching. App.~\ref{sec:hyperparameter_sensitivity} reports sensitivity to the controller targets.

% \paragraph{Online Threshold Adaptation.}
% \textit{Static Thresholds} fixes both routing and switching thresholds at their initial values. This variant remains competitive on several individual benchmarks but reduces the average policy score to 67.9 and RB2 to 47.1. Thus, online calibration provides a modest aggregate improvement, complementing the larger gains from routing and event-triggered switching. Target sensitivity is reported in App.~\ref{sec:hyperparameter_sensitivity}.

% \paragraph{Training Paradigm.}
% \textbf{Mixed-batch joint training} updates the policy and rubric generator on mixed training examples within each batch, removing phase separation and testing the necessity of sequential optimization.
% \textbf{Fixed-length 50/50 alternation} alternates equal-length policy and rubric phases under a predefined schedule, without training-signal routing or event-triggered switching.

% \paragraph{Rubric-Generator Reward.}
% \textbf{Single-pair absolute-margin reward} replaces the full method's scale-normalized $k$-pair reward with an unnormalized margin on one cached pair, isolating pool-level normalization and multi-pair averaging.

\subsection{Training Dynamics}
\label{sec:dynamics}

\begin{figure}[t]
    \centering
    \includegraphics[width=\linewidth]{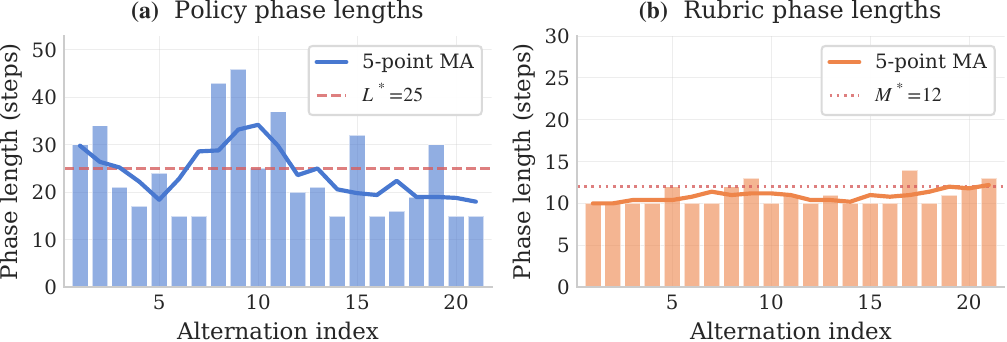}
    \caption{\textbf{Phase-length dynamics.} Policy (a) and rubric (b) phase lengths across successive alternations; solid curves denote 5-point moving averages (MA), and dashed horizontal lines mark the target lengths $L^*=25$ and $M^*=12$, respectively.}
    \label{fig:phase_lengths}
\end{figure}

% \paragraph{Phase-Length Dynamics.}
% Figure~\ref{fig:phase_lengths} reveals a clear stage-dependent pattern in policy phase lengths. After expanding during mid-training and peaking near the 10th alternation, their moving average declines and remains below the target $L^*=25$ in later cycles. Accordingly, the mean length decreases from 27.0 steps in the first half to 21.4 in the second, consistent with reward saturation being reached progressively earlier. In contrast, rubric phases remain stable near $M^*=12$, typically lasting 10--14 steps. Together, these dynamics show that the switching mechanism adapts update allocation to the training state, while keeping rubric recalibration lightweight.

% Figure~\ref{fig:phase_lengths} shows that the policy-phase moving average rises during mid-training and subsequently declines, with the mean phase length decreasing from 27.0 steps in the first half to 21.4 in the second. This suggests that reward saturation is reached earlier as the policy improves, prompting more frequent rubric recalibration. Meanwhile, rubric phases remain compact and close to the target $M^*=12$. Together, these dynamics show that event-triggered switching reallocates optimization effort according to reward informativeness while avoiding unnecessary rubric updates.

Figure~\ref{fig:phase_lengths} shows that the policy-phase moving average rises during mid-training and subsequently declines, with the mean phase length decreasing from 27.0 steps in the first half to 21.4 in the second. This suggests that reward saturation is reached earlier as the policy improves, prompting more frequent rubric recalibration. Meanwhile, rubric phases remain compact and close to the target $M^*=12$. Together, these dynamics show that event-triggered switching reallocates optimization effort according to reward informativeness while avoiding unnecessary rubric updates; per-step model-call counts for policy and rubric updates are detailed in App.~\ref{app:compute_accounting}.

% For per-step online model-call accounting, App.~\ref{app:compute_accounting} reports the policy and rubric update paths separately. With $N_{\mathrm{batch}}=64$ and $G=8$, a policy update uses 512 policy responses, 64 prompt-level rubrics, and 512 judge evaluations, whereas a rubric update reuses cached responses, generates 512 candidate rubrics, and performs 1,024 judge evaluations. These are per-step call counts rather than measured wall-clock or total GPU-hour savings.

% Given the same 500 policy updates, \method uses only 250 rubric updates and performs 750 optimization steps in total. EvoLM uses 500 rubric updates and requires 1,000 total steps. This reduces rubric updates by 50\% and total updates
% by 25\%. Both methods generate one rubric per prompt, reuse it across all $G$ rollouts, and perform $G$ judge evaluations at each policy step. The savings therefore come from avoiding unnecessary rubric updates. \method further requires no proprietary rubric API, task-specific verifier, or external
% preference labels.

%% file: tab/rewardbench.tex
% % \begin{table}[t]
% % \centering
% % \small
% % \setlength{\tabcolsep}{0.5mm}
% % \begin{tabular}{lccccccc}
% % \toprule
% % & \multicolumn{7}{c}{\textbf{RewardBench2}} \\
% % \cmidrule(lr){2-8}
% % \textbf{Method}
% % & \textbf{Factuality}
% % & \textbf{Precise IF}
% % & \textbf{Math}
% % & \textbf{Safety}
% % & \textbf{Focus}
% % & \textbf{Ties}
% % & \textbf{Avg} \\
% % \midrule
% % EvoLM
% % & 37.2 & 30.6 & 54.0 & 43.5 & 54.4 & 56.1 & 46.0 \\
% % RACE
% % & 37.2 & 30.6 & 54.0 & 43.5 & 54.4 & 56.1 & 46.0 \\
% % \bottomrule
% % \end{tabular}
% % \caption{Rubric-quality evaluation on RewardBench2 and JudgeBench.}
% % \label{tab:rewardbench2}
% \begin{table}[t]
% \centering
% \small
% \setlength{\tabcolsep}{0.8mm}
% \begin{tabular}{lcccccc}
% \toprule
% & \multicolumn{6}{c}{\textbf{JudgeBench}} \\
% \cmidrule(lr){2-7}
% \textbf{Method}
% & \textbf{Knowledge}
% & \textbf{Reasoning}
% & \textbf{Math}
% & \textbf{Coding}
% & \textbf{Overall}
% & \textbf{Avg} \\
% \midrule
% EvoLM
% & 31.82 & 36.73 & 26.79 & 33.33 & 32.57 & 32.17 \\
% \textbf{RACE}
% & \textbf{33.12}
% & \textbf{37.76}
% & \textbf{28.57}
% & \textbf{40.48}
% & \textbf{34.57}
% & \textbf{34.98} \\
% \bottomrule
% \end{tabular}
% \begin{tabular}{l cccc}
%   EvoLM   &  \\
%  \textbf{RACE}    & 
% \end{tabular}
% \caption{Evaluation results on JudgeBench and RewardBench 2.}
% \label{tab:judgebench}
% \end{table}

\begin{table}[t]
\centering
\small

% ---------------- JudgeBench ----------------
\resizebox{\linewidth}{!}{
\setlength{\tabcolsep}{0.2mm}
\begin{tabular}{lccccccc}
\toprule
& \multicolumn{6}{c}{\textbf{JudgeBench}}
& \textbf{GPT-5.4} \\
\cmidrule(lr){2-7}
\textbf{Method}
& \textbf{Know.}
& \textbf{Reason.}
& \textbf{Math}
& \textbf{Code}
& \textbf{Overall}
& \textbf{Avg.}
& \textbf{Avg.} \\
\midrule
EvoLM
& 31.8 & 36.7 & 26.8 & 33.3 & 32.6 & 32.2 & 57.9 \\
\textbf{RACE}
& \textbf{33.1}
& \textbf{37.8}
& \textbf{28.6}
& \textbf{40.5}
& \textbf{34.6}
& \textbf{35.0} & \textbf{62.2} \\
\midrule
\end{tabular}}

\vspace{-0.1cm}

\resizebox{\linewidth}{!}{
\setlength{\tabcolsep}{0.3mm}
\begin{tabular}{lcccccccc}
\midrule
& \multicolumn{7}{c}{\textbf{RewardBench2}}
& \textbf{GPT-5.4} \\
\cmidrule(lr){2-8}
\textbf{Method}
& \textbf{Fact.}
& \textbf{IF}
& \textbf{Math}
& \textbf{Safety}
& \textbf{Focus}
& \textbf{Ties}
& \textbf{Avg.}
& \textbf{Avg.} \\
\midrule
EvoLM
& 37.2
& 30.6
& 54.0
& \textbf{43.5}
& 54.4
& 56.1
& 46.0 & 63.4 \\

\textbf{RACE}
& \textbf{42.9}
& \textbf{34.3}
& \textbf{60.2}
& 41.6
& \textbf{58.3}
& \textbf{58.1}
& \textbf{49.2}
& \textbf{65.7} \\
\bottomrule
\end{tabular}}

\caption{JudgeBench and RewardBench2 evaluations.}
\label{tab:judge_rewardbench}
\vspace{-0.3cm}
\end{table}

%% file: tab/ablation.tex
\begin{table}[t]
\centering
\small
\resizebox{\linewidth}{!}{
\setlength{\tabcolsep}{0.5mm}
\begin{tabular}{llcc}
\toprule
\textbf{Dimension}
& \textbf{Variant}
& \textbf{RB2}
& \textbf{Policy Avg.} \\
\midrule

Overall
& \method (full)
&  \textbf{49.2}
& \textbf{73.0} \\

\midrule
\multirow{3}{*}{Routing}
& No Gradient Filtering
& \underline{48.7}
& 70.8 \\
& Uniform Rubric Sampling
& 47.9
& 71.4 \\
& Filter-only
& 45.1
& 70.9 \\

\midrule
\multirow{2}{*}{Switching}
& Random Phase-Length Alternation
& 45.3
& 71.4 \\
& Fixed 50/50 Alternation + Routing
& 44.3
& 71.3 \\

\midrule
Adaptation
& Static Thresholds
& 47.1
& \underline{72.1} \\

\bottomrule
\end{tabular}}
\caption{\textbf{Ablation study.}
RB2 and Policy Avg.\ denote the average performance on RewardBench2 and the eight main-table downstream benchmarks, respectively. \textbf{Bold} and \underline{underlined} values indicate the best and second-best results in each column.}
\label{tab:ablations}
\vspace{-0.4cm}
\end{table}

%% file: sec/conclusions.tex
%% ============================================================
\section{Conclusion}
\label{sec:conclusion}
%% ============================================================

% We presented \method, an event-triggered controller for policy-rubric co-training that detects reward saturation and adaptively allocates training budget. EMA-based phase switching and per-rollout-group routing produce fully data-driven training dynamics, with all adaptive thresholds governed by a unified proportional control mechanism that empirically tracks its target rates and replaces fragile hand-tuned thresholds with stable intent-expressing targets (only 4 tunable parameters). The routing mechanism serves a dual role: it filters uninformative gradients from the policy update while simultaneously collecting high-value training data for the rubric phase, creating an information pathway from failure identification to targeted retraining. Across 10 benchmarks spanning reasoning, code, and knowledge, \method achieves consistent gains on hard reasoning tasks over fixed-schedule baselines.

We identify heterogeneous reward-discriminability degradation as a key bottleneck in policy--rubric co-evolution: some rollout groups experience advantage collapse while others remain informative. To address this, we introduce \method, combining group-level signal routing, event-triggered phase switching, and online threshold adaptation. \method achieves the highest average performance and outperforms the strongest fixed-alternation baseline. These gains are achieved with a single frozen judge, without task-specific verifiers or preference-labeled online rewards, and with fewer optimization updates than fixed-alternation training. Overall, these results demonstrate the effectiveness of reward-adaptive policy--rubric co-evolution.

%% file: sec/limitation.tex
\section*{Limitations}
Due to computational constraints, our experiments focus on a shared Qwen3-8B policy--rubric backbone trained with a frozen Qwen3-1.7B judge. Although we additionally evaluate rubric discrimination with another judge, the generalizability of our training framework to other model families, scales, and judges remains to be explored. 

Finally, our downstream evaluation focuses on single-turn benchmarks covering reasoning, knowledge, and code generation. Extending reward-adaptive co-evolution to multilingual, multi-turn, and long-horizon settings is a promising direction for future work.

%% file: sec/app.tex
\section{Algorithm}
\label{app:algorithm}

The complete training procedure is given in Algorithm~\ref{alg:tempo}.

\input{tab/algorithm}

\section{Implementation Details}
\label{app:implementation}

\subsection{Model Configuration}

We use Qwen3-8B (8.2B parameters) for both the policy and the rubric generator. The two roles share a single parameter set and are distinguished by role-specific chat templates; consequently, policy and rubric updates are applied sequentially and never overwrite the shared weights concurrently. A frozen Qwen3-1.7B model supplies all rubric-conditioned scores.

\subsection{Optimization and Sampling}

Both the policy and rubric generator are optimized using GRPO. We adopt a learning rate of $10^{-6}$ and a KL coefficient of $\beta=10^{-3}$, with one optimization epoch and one minibatch per update. The learning rate is held constant without warmup. We use the \texttt{kl3} estimator, a per-device microbatch size of 1, and a fused optimizer.

Each update contains $N_{\mathrm{batch}}=64$ unique prompts, with $G=8$ samples per prompt, yielding 512 training samples. Actor sampling uses a temperature of 1.0. Prompts are truncated to 2,048 tokens, while generations and packed sequences are capped at 16,384 and 18,500 tokens, respectively. The asynchronous rollout pipeline supports a lag of up to four steps and caps the truncated importance-sampling ratio at 2.0.  We report results for \method across three independent runs and use previously reported baseline results.

\subsection{Full Hyperparameter Table}
\label{app:hyperparams}

For completeness, Table~\ref{app:diagnostics} reports the full optimization, sampling, infrastructure, and controller configuration used in our experiments.

\begin{table*}[t]
\begin{center}
\small
\begin{tabular}{lll}
\toprule
\textbf{Parameter} & \textbf{Value} & \textbf{Description} \\
\midrule

\multicolumn{3}{l}{\textit{Training Infrastructure}} \\
Batch size (unique prompts) & 64 & Per policy step \\
Rollouts per prompt ($G$) & 8 & For GRPO \\
Max response length & 16384 & Tokens \\
GPUs & 16 & 2 nodes $\times$ 8 H800 \\
Learner ranks & 8 & 4 per node, DeepSpeed ZeRO-3 \\
vLLM engines & 8 & Tensor parallelism 1 \\
\midrule
% \multicolumn{3}{l}{\textit{Phase switching}} \\
% $t_{\min}^{\text{policy}}$ & 15 & Minimum policy steps before trigger \\
% $t_{\max}^{\text{policy}}$ & 50 & Hard cap on policy phase length \\
% $t_{\min}^{\text{rubric}}$ & 12 & Minimum rubric steps before trigger \\
% $t_{\max}^{\text{rubric}}$ & 30 & Hard cap on rubric phase length \\
% \midrule
\multicolumn{3}{l}{\textit{Per-Rollout-Group Training-Signal Routing}} \\
$\tau_{\mathrm{tie}}^\star$ & 0.70 & Initial tie-rate threshold \\
$\sigma_{\mathrm{rt}}$ & 0.03 & Initial routing std threshold \\
$d^\star$ & 0.05 & Distinct-trigram floor for diversity collapse \\
\midrule
\multicolumn{3}{l}{\textit{Policy $\to$ Rubric: Reward-Saturation Trigger}} \\
$\sigma_{\mathrm{sw}}^\star$ & 0.06 & Reward std threshold for phase switching \\
% $[\mu_{\text{lo}}, \mu_{\text{hi}}]$ & $[0.5, 1.01]$ & Mean reward range filter \\
$\tau_{\mathrm{sat}}^{\mathrm P}$ & 0.40 & EMA threshold for trigger (initial; adapted online) \\
% $S$ & 5 & EMA span \\
% patience & 3 & Consecutive steps above $\varphi$ \\
\midrule
\multicolumn{3}{l}{\textit{Rubric $\to$ Policy: Rubric-Improvement Plateau Trigger}} \\
Window $W$ & 5 & Rolling comparison window \\
$\tau_{\mathrm{imp}}^{\mathrm{R}}$ & 0.005 & Rubric plateau threshold (initial; adapted online) \\
% $\epsilon_{\text{abs}}^{\text{policy}}$ & 0.01 & Policy plateau threshold \\
% $\epsilon_{\text{rel}}$ & 0.15 & Relative gain threshold (level metrics) \\
% patience & 2 & Consecutive low-gain windows \\
% cooldown & 8 & Minimum steps before plateau checks \\
\midrule
\multicolumn{3}{l}{\textit{Online Threshold Adaptation: routing (TRQC)}} \\
$q^*$ & 0.20 & Target quarantine rate \\
$\alpha_q$ & 0.1 & EMA smoothing for quarantine rate \\
$\eta$ & 0.003 & Proportional step size \\
% $[\sigma_{\mathrm{rt}}^{\min}, \sigma_{\mathrm{rt}}^{\max}]$ & $[0.01, 0.10]$ & Projection bounds on $\sigma_{\mathrm{rt}}$ \\
% $[\tau_{\text{tie}}^{\min}, \tau_{\text{tie}}^{\max}]$ & $[0.40, 0.90]$ & Projection bounds on $\tau_{\text{tie}}$ \\
\midrule
\multicolumn{3}{l}{\textit{Online Threshold Adaptation: switching}} \\
$L^*$ & 25 & Target policy-phase length (steps) \\
% $\delta_\varphi$ & 0.02 & Proportional step size for $\varphi$ ($2\delta_\varphi$ on hard-cap exit) \\
% $W_\varphi$ & 5 & Cycle window ($\geq$3 trigger-terminated phases required) \\
% $[\varphi_{\min}, \varphi_{\max}]$ & $[0.20, 0.70]$ & Projection bounds on $\varphi$ \\
$M^*$ & 12 & Target rubric-phase length (steps) \\
% $\delta_{\mathrm R}$ & 0.02 & Proportional rate for rubric threshold \\
% $[\epsilon_{\mathrm R}^{\min},\epsilon_{\mathrm R}^{\max}]$ & $[0.001,0.05]$ & Rubric-threshold bounds \\
\midrule
\multicolumn{3}{l}{\textit{Rubric Generator Optimization}} \\
% $k$ & 1 & Cached response pairs per candidate rubric \\
% $m_{\mathrm{ref}}$ fallback & 0.2 & Reward normalization fallback \\
$c$ (clip) & 1.0 & Discrimination-reward clipping value \\
% Format weight & 0.3 & JSON-format reward mixture weight \\
Rubric training pool $\mathcal{P}$ size & 2048 & Rubric training pool capacity \\
\bottomrule
\end{tabular}
\end{center}
\caption{Complete hyperparameter specification for \method.}
\label{app:diagnostics}
\end{table*}

% Both uninformative conditions share the reward-side trigger: threshold comparisons on reward standard deviation and tie rate (Eq. 8). The diversity gate then attributes the failure. A group whose outputs are homogeneous ($\operatorname{cv}_{\ell} \leq \operatorname{cv}^\star = 0.2$ and $d_{\mathrm{tri}} \leq d^\star = 0.05$) is classified as diversity-collapsed: near-identical rollouts \textbf{should} receive similar rewards. Hence, the tie is not the rubric's fault, and the prompt is excluded from the policy update without being added to $\mathcal{P}$. Only groups whose rollouts remain diverse are classified as rubric-failed and routed to $\mathcal{P}$.

% \section{Trained Rubrics across Judge Models}

\section{Full RewardBench2 Evaluation}
\label{app:full_rewardbench}

Table~\ref{tab:rewardbench2_full} reports the full RewardBench2 results. Compared with EvoLM, \method improves five of the six dimensions and raises the average score from 46.0 to 49.2. It also outperforms both prompted-rubric baselines and RLCER.

The scalar reward model presents a notable contrast: despite achieving the highest RewardBench2 score, it yields the weakest downstream policy performance among the compared methods, trailing \method by 9.5 points. Similarly, Rubric-ARM obtains the strongest RewardBench2 result among the co-evolution methods but underperforms \method on downstream policy evaluation. These results suggest that discrimination on a static reward-model benchmark does not necessarily translate into effective supervision for an evolving policy. In contrast, \method improves both rubric discrimination relative to EvoLM and downstream policy performance, supporting the benefit of adapting rubric supervision to the current policy distribution.

\input{tab/rewardbenchfull}

\section{Rubric Examples}
\label{app:rubric_examples}

We analyze rubrics generated for 100 fixed prompts spanning mathematical reasoning, code generation, constrained writing, and scientific explanation. For a controlled comparison, we evaluate both methods after 500 policy updates, corresponding to step 700 for \method and step 1000 for EvoLM. At these checkpoints, \method has completed 200 rubric updates, compared with 500 under EvoLM's fixed alternation. Thus, the comparison holds policy updates constant while \method uses substantially fewer rubric updates and total optimization steps.

\subsection{Same-Prompt Comparison}

\begin{table*}[t]
\centering
\small
\begin{tabular}{@{}l c p{0.76\textwidth}@{}}
\toprule
\textbf{Generator} & \textbf{\#} & \textbf{Criteria emitted for the rectangle prompt} \\
\midrule
EvoLM, step 1000 & 4 & Relate length and width using the perimeter formula; express area in one variable; identify a maximization method; provide the correct maximum-area value. \\
RACE, step 700 & 5 & Set up $2x+2y=48$ and $A=xy$; apply a valid maximization method; explain why the optimum is a square; give 144 square units; present the solution with a clear logical flow. \\
\bottomrule
\end{tabular}
\caption{Same-prompt rubric comparison. The descriptions preserve the semantic content of the generated criteria while removing only the repeated three-level scoring anchors for compactness.}
\label{tab:rubric_example_math_compare}
\end{table*}

\paragraph{Math problem.}
Consider the prompt: \emph{``The perimeter of a rectangle is 48. What is the
largest possible area of the rectangle?''}
Table~\ref{tab:rubric_example_math_compare} compares the rubrics generated
for this prompt. EvoLM describes the solution through broadly applicable
stages, whereas \method grounds the criteria in the specific problem. In particular, it explicitly requires identifying the square as the maximizing
rectangle and reporting the maximum area of 144 square units. It also separates answer correctness from explanation quality, allowing the judge to
distinguish a correct numerical answer from a complete derivation. This
example illustrates how \method converts generic procedural criteria into
more specific and verifiable checks.

\subsection{Rubric Evolution over Training}

% For the same rectangle prompt, Table~\ref{tab:rubric_example_math_evolution} tracks RACE at three checkpoints. The number of criteria does not grow monotonically. Instead, the content moves from an outcome-heavy checklist at step 50 toward task-specific intermediate equations at step 350, before being repartitioned into setup, optimization, geometric justification, final value, and presentation at step 700. In particular, the step-350 rubric names the substitution $y=24-x$ and the quadratic $A=24x-x^2$, making partial-credit decisions more directly checkable by the frozen judge.

For the same rectangle prompt, Table~\ref{tab:rubric_example_math_evolution} tracks RACE at three checkpoints. The number of criteria does not grow monotonically. Instead, the content moves from an outcome-heavy checklist at step 50 toward task-specific intermediate equations at step 350, before being repartitioned into setup, optimization, geometric justification, final value, and presentation at step 700. In particular, the step-350 rubric names the substitution $y=24-x$ and the quadratic $A=24x-x^2$, making partial-credit decisions more checkable by the frozen judge.

\begin{table*}[t]
\centering
\small

\begin{tabular}{@{}c c p{0.80\textwidth}@{}}
\toprule
\textbf{Step} & \textbf{\#} & \textbf{Representative criteria} \\
\midrule
50 & 5 & Correct value 144; derive the perimeter relation; use an appropriate maximization method; explain why the optimum is a square; maintain clarity and logical flow. \\
350 & 4 & Set up perimeter and area equations; substitute $y=24-x$ to obtain $A=24x-x^2$; identify a valid maximization argument; give 144 with justification. \\
700 & 5 & Set up perimeter and area; maximize the area; justify the square optimum; give 144 square units; show a clear progression from setup to solution. \\
\bottomrule
\end{tabular}
\caption{RACE rubric evolution for one fixed math prompt. Criteria are abbreviated, while retaining the emitted equations, expected values, and distinctions between criteria.}
\label{tab:rubric_example_math_evolution}
\end{table*}

\subsection{Aggregate Rubric Statistics}

Table~\ref{tab:rubric_example_aggregate} summarizes rubric characteristics across 100 fixed prompts. From step 50 to 700, JSON parse validity increases from 73\% to 90\%, despite fluctuations at intermediate steps. Among parse-valid rubrics, the proportion of label-only criteria decreases from 20.9\% to 9.9\%, while criteria containing concrete expected values increase from 10.1\% to 18.0\%. Mean criterion length rises from 56.9 to 62.6 characters, with roughly five criteria per rubric. Compared with EvoLM at step 1000, \method at step 700 achieves higher parse validity (90\% vs. 82\%) and higher specificity (18.0\% vs. 14.5\%), fewer label-only criteria (9.9\% vs. 11.8\%), and longer criteria on average (62.6 vs. 59.8 characters).

\begin{table*}[t]
\centering
\small
\resizebox{\linewidth}{!}{\begin{tabular}{lccccccc}
\toprule
\textbf{Generator} & \textbf{Valid (\%)} & \textbf{\# Crit.} & \textbf{Len.} & \textbf{Label-only (\%)} & \textbf{Specific (\%)} & \textbf{Constraint (\%)} & \textbf{Grounded (\%)} \\
\midrule
RACE, step 50  & 73 & 5.5 & 56.9 & 20.9 & 10.1 & 10.8 & 38.1 \\
RACE, step 350 & 79 & 4.9 & 63.0 & 10.4 & 14.7 & 11.9 & 43.8 \\
RACE, step 700 & 90 & 5.0 & 62.6 &  9.9 & 18.0 & 11.6 & 40.0 \\
EvoLM, step 1000 & 82 & 5.2 & 59.8 & 11.8 & 14.5 & 10.6 & -- \\
\bottomrule
\end{tabular}}

\caption{Rubric statistics on 100 fixed prompts. Validity is the percentage of prompts yielding parseable JSON. The remaining metrics are computed over criteria from parse-valid rubrics. ``Specific'' denotes criteria containing concrete expected values; ``Grounded'' denotes lexical overlap with prompt-specific entities or constraints.}
\label{tab:rubric_example_aggregate}
\end{table*}

These statistics characterize rubric form and lexical content rather than directly measuring reward discrimination. Nevertheless, they suggest that \method produces more task-specific criteria with fewer underspecified labels.

\subsection{Scale-normalized Pairwise Reward Diagnostics}
\label{app:rubric_reward_diagnostics}

We examine the scale-normalized pairwise reward $S_{\mathrm{rubric}}$ used to optimize the rubric generator, following Eq.~\ref{eq:rubric_train_reward} with clipping value $c=1.0$. For each candidate rubric, the frozen judge evaluates the same cached response pair, producing one candidate--pair reward. Candidate rubrics evaluated for the same training group form a rubric group; variation among their final rewards is the signal relevant to the group-level GRPO update. The analysis covers rubric groups from rubric training phase steps 31--700. These counts specify the statistical coverage and denominators of the diagnostics summarized in Table~\ref{tab:rubric_reward_diagnostics_summary}.

\begin{table}[t]
\centering
\small

\begin{tabular}{@{}l r}
\toprule
\textbf{Diagnostic} & \textbf{Value}  \\
\midrule
near tie ($|m_{old}| < 0.025$) & 5.23\%\\
Advantage collapse ($\sigma(S_{\mathrm{rubric}})=0$) & 4.56\%   \\
Base reward at clipping boundary & 12.19\%  \\
\bottomrule
\end{tabular}%

\caption{Rubric-generator reward diagnostics.}
\label{tab:rubric_reward_diagnostics_summary}
\end{table}

Only 4.56\% of the retained groups satisfy the advantage-collapse condition ($\sigma(S_{\mathrm{rubric}})=0$), leaving 95.44\% with nonzero within-group reward variation. This indicates that the rubric-phase reward is not broadly degenerate in the retained subsample and remains usable for most groups.

\subsection{Policy-Phase Reward Saturation}
\label{app:delta_analysis}

The statistics above characterize rubric form. We now measure the quantity that actually drives policy learning: whether the current rubric still separates the responses the current policy produces. Because GRPO centers advantages within each rollout group, a group whose $G$ responses all receive the same rubric score contributes no policy gradient no matter how high that score is. The failure mode we must track is therefore loss of \emph{within-group spread}, not low mean reward.

% \paragraph{Group-level view.} Table~\ref{tab:policy_saturation} reports, over the rollout groups logged during \method's policy phases, the fraction with zero score spread and the mean within-group standard deviation. Zero-spread groups account for $24.5\%$ of all groups and the rate does not drift upward across training ($24.2\%$ at step 50, $22.4\%$ at step 700), but the mean spread does contract steadily ($0.073$ to $0.064$) while the fraction of groups in which every response scores at least $0.8$ holds near $75\%$. The reward is thus becoming progressively harder to use even where it has not collapsed outright.

% \paragraph{Group-level view.} Table~\ref{tab:policy_saturation} reports, over the rollout groups during \method's policy phases, the fraction satisfying the advantage-collapse condition ($\sigma(\mathbf r)=0$) on the retained subsamples and the mean within-group standard deviation. This condition holds for $24.5\%$ of the evaluated groups and the rate does not drift upward across training ($24.2\%$ at step 50, $22.4\%$ at step 700), but the mean spread does contract steadily ($0.073$ to $0.064$) while the fraction of groups in which every response scores at least $0.8$ holds near $75\%$. The reward is thus becoming progressively harder to use even where it has not collapsed outright.

\paragraph{Group-level view.} Table~\ref{tab:policy_saturation} reports, over the rollout groups evaluated during \method's policy phases, the fraction satisfying the advantage-collapse condition ($\sigma(\mathbf r)=0$) on the retained subsamples and the mean within-group standard deviation. This condition holds for $24.5\%$ of the evaluated groups and the rate does not drift upward across training ($24.2\%$ at step 50, $22.4\%$ at step 700), but the mean spread does contract steadily ($0.073$ to $0.064$) while the fraction of groups in which every response scores at least $0.8$ holds near $75\%$. The reward is thus becoming progressively harder to use even where it has not collapsed outright.

\begin{table}[t]
\centering
\small
\resizebox{\linewidth}{!}{
\begin{tabular}{@{}c r c c c@{}}
\toprule
\textbf{Step} & \textbf{Groups} & \textbf{$\sigma(\mathbf r)=0$ (\%)} & \textbf{All $\geq 0.8$ (\%)} & \textbf{Mean $\sigma(\mathbf r)$} \\
\midrule
50  & 4{,}684  & 24.2 & 74.4 & 0.073 \\
100 & 2{,}970  & 28.5 & 75.3 & 0.070 \\
200 & 3{,}547  & 28.6 & 75.8 & 0.066 \\
250 & 4{,}017  & 23.9 & 75.1 & 0.069 \\
350 & 6{,}833  & 22.8 & 75.2 & 0.068 \\
500 & 5{,}718  & 24.6 & 76.2 & 0.065 \\
700 & 4{,}884  & 22.4 & 75.8 & 0.064 \\
\midrule
All & 32{,}653 & 24.5 & 75.4 & 0.068 \\
\bottomrule
\end{tabular}}
\caption{Reward variation within policy rollout groups during \method training. Each checkpoint row aggregates groups from a disjoint window centered on that checkpoint.}
\label{tab:policy_saturation}
\end{table}

\input{tab/reward_computation}

\input{tab/call}

\input{tab/hyperparameter}

\paragraph{Saturation, not rubric failure.} We next decompose the groups satisfying the advantage-collapse condition by absolute score level. $95.5\%$ of the groups with $\sigma(\mathbf r)=0$ have every response scored at or above $0.8$ and only $2.2\%$ have every response at or below $0.2$. Collapse is therefore almost never caused by a rubric that rejects everything; it is caused by a rubric whose criteria are all satisfied by every response the policy now produces. This is precisely the mechanism identified in Section~\ref{sec:observation}: as the policy improves, a rubric held fixed loses discriminative power on responses that already meet it, so the reward saturates from above.

\section{Per-Step Compute Accounting}
\label{app:compute_accounting}

We report online compute per optimizer update using the symbols of
Section~\ref{sec:switching}: $N_{\mathrm{batch}}$ prompt groups per update and $G$ GRPO samples per group, which are policy responses $\{y_i\}_{i=1}^{G}$ during a policy update and candidate rubrics $\{R_i\}_{i=1}^{G}$ during a rubric update. A single frozen judge $\mathcal J$ produces every rubric-conditioned score $J(x,y;R)$. Our run uses $(N_{\mathrm{batch}},G)=(64,8)$. We count autoregressive actor generations and judge evaluations separately because their
sequence lengths and model sizes differ. Data loading, pool sampling, margin normalization, routing statistics, JSON-format checks, and controller updates require no additional model inference. Offline costs are excluded. Table~\ref{tab:compute_accounting} compares the reward-side work with the baselines following the accounting convention of EvoLM, while Table~\ref{tab:race_step_compute} expands the two update types used by RACE.

RRD has data-dependent construction cost because recursive refinement expands and filters candidates until convergence. EvoLM and the RACE policy update reuse one prompt-level rubric across all $G$ responses. RACE additionally trains the
rubric generator: this update reuses cached policy responses, but each of the
$G$ candidate rubrics requires scoring both members of one cached pair and
normalizes the resulting margin by the cached historical margin
$m_{\mathrm{old}}$ in Eq.~\ref{eq:old_margin}.

For a policy update, the shared Qwen3-8B model generates
$N_{\mathrm{batch}}G=512$ responses and $N_{\mathrm{batch}}=64$
prompt-level rubrics; the frozen Qwen3-1.7B judge performs
$N_{\mathrm{batch}}G=512$ rubric-conditioned answer evaluations. The learner then performs one GRPO epoch using the informative groups selected by routing. Thus, up to variable
sequence lengths, the online cost can be written as
\begin{align*}
\mathcal C_{\mathrm P}
&= N_{\mathrm{batch}}G\mathcal C_{\pi}^{\mathrm{gen}}
+ N_{\mathrm{batch}}\mathcal C_{\mathcal G}^{\mathrm{gen}} \\
&\quad + N_{\mathrm{batch}}G\mathcal C_{\mathcal J}^{\mathrm{eval}}
+ \mathcal C_{\mathrm{GRPO}}(N_{\mathrm{batch}}G).
\end{align*}

For a rubric update, the policy responses are read from the rubric training pool $\mathcal P$ and are not regenerated. The shared model generates $N_{\mathrm{batch}}\times G=512$ candidate rubrics. Every candidate is scored on both members of its cached ordered pair $(y^+,y^-)$, requiring $2N_{\mathrm{batch}}G=1{,}024$ judge evaluations, followed by one GRPO epoch over 512 rubric samples:
\begin{equation}
\begin{aligned}
\mathcal C_{\mathrm R}
&= N_{\mathrm{batch}}G\mathcal C_{\mathcal G}^{\mathrm{gen}}
+ 2N_{\mathrm{batch}}G\mathcal C_{\mathcal J}^{\mathrm{eval}}\\
&\quad + \mathcal C_{\mathrm{GRPO}}(N_{\mathrm{batch}}G).
\end{aligned}
\end{equation}
The margins $m_i$ of Eq.~\ref{eq:new_margin} are differences of judge scores that are already counted above, and the cached anchor $m_{\mathrm{old}}$ of Eq.~\ref{eq:old_margin} was computed when the pair entered the rubric training pool $\mathcal P$ during a policy phase. The routing statistics and all TRQC and phase-switching statistics are likewise computed from cached scalar scores and generated texts. Routing and the pairwise-margin reward therefore add no model calls beyond the counts above.

\section{Hyperparameter Sensitivity}
\label{sec:hyperparameter_sensitivity}

We study the sensitivity of \method to the three controller targets:
the quarantine target $q^*$ in TRQC and the target policy- and rubric-phase lengths $L^*$ and $M^*$ in the adaptive phase-switching controller. The
default configuration uses $q^*=0.20$, $L^*=25$, and $M^*=12$. We vary $q^*$ while keeping the phase-length targets fixed, and jointly vary $(L^*, M^*)$ to examine different alternation timescales. Specifically, we compare the default setting with a short-phase configuration $(L^*, M^*)=(10,5)$. Notably, $L^*$ and $M^*$ are soft operating points
used to calibrate the switching thresholds; actual transitions remain determined by reward saturation and rubric-reward improvement.

Table~\ref{tab:hyperparameter_sensitivity} reports downstream policy performance under these settings. A lower $q^*$ reduces the fraction of quarantined rollout groups and retains more policy supervision, whereas a higher target permits more aggressive filtering. Shorter phase-length targets encourage more frequent policy--rubric alternation.
This comparison evaluates whether \method depends on a particular quarantine rate or alternation timescale.

\section{IFEval Evaluation Protocol}
\label{app:ifeval}

IFEval is reported separately from the main downstream table because the default evaluation protocol applies constraint checks to reasoning text as well as the final answer. Following EvoLM, we use the \texttt{olmo3:adapt} suite from the OLMo evaluation framework, whose legacy implementation is available in the \href{https://github.com/allenai/OLMo-Eval}{OLMo-Eval repository}. Under this configuration, constraint checkers operate on the full raw Qwen3 completion, including the \texttt{<think>...</think>} block. Consequently, internal reasoning may trigger constraint violations unrelated to the final answer.

For a controlled comparison, we re-evaluate reproduced EvoLM and \method with thinking enabled during generation, but extract and score only the final answer after removing the \texttt{<think>...</think>} block. Under this final-answer-only protocol, EvoLM and \method score 77.6 and 78.2, respectively. These scores are directly comparable to each other, but not to EvoLM's paper-reported IFEval score obtained under the full-completion protocol.

\section{The Use of Large Language Models}
We used a large language model (LLM) solely to improve linguistic clarity and sentence structure in text based on ideas and analyses developed by the human authors. The LLM did not contribute to research ideation, experimental design, data analysis, or scientific conclusions.

%% file: tab/algorithm.tex
\begin{algorithm}[H]
\caption{\method}
\label{alg:tempo}
\small
\begin{algorithmic}[1]
\REQUIRE Dataset $\mathcal{D}$, shared weight $\theta$, judge $\mathcal{J}$, group size $G$
\STATE $s\gets\texttt{policy}$;
rubric training pool $\mathcal{P}\gets\emptyset$;
$\hat f_{\mathrm{sat}}\gets0$; $t_R\gets0$

\WHILE{training budget remains}

\IF{$s=\texttt{policy}$}
    \STATE Sample prompts
    $\{x_j\}_{j=1}^{N_{\mathrm{batch}}}\sim\mathcal{D}$
    \STATE Initialize informative groups $\mathcal{I}\gets\emptyset$

    \FOR{$j=1,\ldots,N_{\mathrm{batch}}$}
        \STATE Generate
        $R_j\sim\mathcal{G}_\theta(\cdot\mid x_j)$ and
        $\{y_{j,i}\}_{i=1}^{G}\sim\pi_\theta(\cdot\mid x_j)$
        \STATE Compute
        $r_{j,i}\gets S_{\mathrm{policy}}(x_j,y_{j,i};R_j)$
        for all $i$
        \STATE Route group $j$ using
        Eqs.~\ref{eq:reward_degenerate}--%
        \ref{eq:surface_form_collapse}

        \IF{group $j$ is informative}
            \STATE Add group $j$ to $\mathcal{I}$

        \ELSIF{group $j$ is rubric-failed}
            \STATE
            $i_j^+\gets\arg\max_i r_{j,i}$;
            $i_j^-\gets\arg\min_{i\neq i_j^+}r_{j,i}$
            \STATE
            $y_j^+\gets y_{j,i_j^+}$;
            $y_j^-\gets y_{j,i_j^-}$;
            $m_{\mathrm{old},j}\gets
            r_{j,i_j^+}-r_{j,i_j^-}$
            \STATE Push
            $(x_j,y_j^+,y_j^-,m_{\mathrm{old},j})$
            into $\mathcal{P}$
        \ENDIF
    \ENDFOR

    \STATE Update $\pi_\theta$ by GRPO using groups in $\mathcal{I}$
    \STATE Update TRQC using
    Eqs.~\ref{eq:trqc_rate}--\ref{eq:trqc_update}
    \STATE Compute $f_{\mathrm{sat}}$ and update
    $\hat f_{\mathrm{sat}}$ using
    Eqs.~\ref{eq:reward_saturation_rate}--%
    \ref{eq:saturation_ema}

    \IF{$\hat f_{\mathrm{sat}}\geq\tau_{\mathrm{sat}}^{\mathrm P}$}
    \STATE Update $\tau_{\mathrm{sat}}^{\mathrm P}$ using
    Eqs.~\ref{eq:phase_length_errors}--\ref{eq:adaptive_phase_thresholds}
    \STATE $s\gets\texttt{rubric}$; $t_R\gets0$
    \ENDIF

\ELSE
    \STATE $t_R\gets t_R+1$
     \STATE $\mathcal E=
    \{(x_j,y_j^+,y_j^-,m_{\mathrm{old},j})\}
    _{j=1}^{N_{\mathrm{batch}}}\subseteq\mathcal P$

    \STATE For each entry in $\mathcal E$, generate $G$ candidate rubrics
    and compute their rewards using
    Eqs.~\ref{eq:new_margin}--\ref{eq:rubric_train_reward}

    \STATE Update $\mathcal G_\theta$ by GRPO

    \IF{$t_R\geq2W$}
        \STATE Compute $\Delta_W^{t_R}$ using
        Eq.~\ref{eq:rubric_plateau}

        \IF{$\Delta_W^{(t_R)}
            \leq\tau_{\mathrm{imp}}^{\mathrm R}$}
            \STATE Record $t_R$ and update
            $\tau_{\mathrm{imp}}^{\mathrm R}$ using
            Eqs.~\ref{eq:phase_length_errors}--%
            \ref{eq:adaptive_phase_thresholds}
            \STATE $s\gets\texttt{policy}$;
            $t_P\gets0$;
            $\hat f_{\mathrm{sat}}\gets0$
        \ENDIF
\ENDIF
\ENDIF

\ENDWHILE
\end{algorithmic}
\end{algorithm}

%% file: tab/rewardbenchfull.tex
 \begin{table*}[t]
\centering
\small
\setlength{\tabcolsep}{3.2pt}
\renewcommand{\arraystretch}{1.05}
\begin{tabular*}{\textwidth}{
@{\extracolsep{\fill}}lccccccc@{}
}
\toprule
\textbf{Method}
& \textbf{Factuality}
& \textbf{Precise IF}
& \textbf{Math}
& \textbf{Safety}
& \textbf{Focus}
& \textbf{Ties}
& \textbf{Avg.} \\
\midrule

GPT-4.1 (prompted)
& 32.1
& 32.6
& 34.4
& 39.1
& 35.0
& 46.3
& 36.6 \\

Qwen3-8B (prompted)
& 27.8
& 26.5
& 38.1
& 25.0
& 30.2
& 9.7
& 26.2 \\

Scalar RM (Skywork-RM-V2)
& \textbf{88.2}
& \textbf{67.8}
& \textbf{83.1}
& \textbf{97.3}
& \textbf{99.2}
& \textbf{83.1}
& \textbf{86.4} \\

\midrule

RaR
& \multicolumn{7}{c}{N/A (frozen GPT-4.1 rubric)} \\

RRD
& \multicolumn{7}{c}{N/A (frozen GPT-4.1 rubric)} \\

RLCER
& 33.1
& 29.8
& 52.2
& 48.3
& 51.9
& 54.8
& 45.0 \\

Rubric-ARM
& \underline{48.8}
& \underline{39.1}
& \underline{63.5}
& \underline{60.8}
& \underline{66.1}
& \underline{58.2}
& \underline{56.1} \\

EvoLM
& 37.2
& 30.6
& 54.0
& 43.5
& 54.4
& 56.1
& 46.0 \\

\method
& 42.9
& 34.3
& 60.2
& 41.6
& 58.3
& 58.1
& 49.2 \\

\bottomrule
\end{tabular*}
\caption{\textbf{Full RewardBench2 evaluation.}
All scores are percentages. \textbf{Bold} and \underline{underlined} values
indicate the best and second-best results in each column, respectively.
N/A denotes methods for which a directly comparable trained rubric generator
is unavailable.}
\label{tab:rewardbench2_full}
\end{table*}

%% file: tab/reward_computation.tex
\begin{table*}[t]
\centering
\small
\begin{tabular}{@{}l p{0.21\textwidth} p{0.20\textwidth} p{0.30\textwidth}@{}}
\toprule
\textbf{Method}
& \textbf{Rubric construction}
& \textbf{Answer scoring}
& \textbf{Extra checks} \\
\midrule
RaR
& 1 rubric proposal
& $G$ answer scores
& -- \\
RRD
& Initial proposal + recursive refinement
& $G \times I$ item-level scores
& Data-dependent refinement checks \\
RLCER
& $G$ rubric generations (one per rollout)
& $G^2$ judge calls
& $G$ verifier checks \\
Rubric-ARM
& 1 rubric generation (frozen model)
& $G$ answer scores
& -- \\
EvoLM
& 1 prompt-level rubric generation
& $G$ answer scores
& -- \\
RACE (policy update)
& 1 prompt-level rubric generation
& $G$ answer scores
& -- \\
RACE (rubric update)
& $G$ candidate rubric generations
& $2G$ cached-pair answer scores
& Pool lookup (no model calls) \\
\bottomrule
\end{tabular}
\caption{Per-prompt online reward computation. Here $G$ is the GRPO group size and $I$ denotes the number of rubric items scored individually by the corresponding baseline.}
\label{tab:compute_accounting}
\end{table*}

%% file: tab/call.tex
\begin{table*}[t]
\centering
\small
\begin{tabular}{lllll}
\toprule
\textbf{Update}
& \textbf{Policy generations}
& \textbf{Rubric generations}
& \textbf{Judge evaluations}
& \textbf{Learner samples} \\
\midrule
Policy
& $N_{\mathrm{batch}}G$
& $N_{\mathrm{batch}}$
& $N_{\mathrm{batch}}G$
& $N_{\mathrm{batch}}G$ policy responses \\
Rubric
& 0 (cached responses)
& $N_{\mathrm{batch}}G$
& $2N_{\mathrm{batch}}G$
& $N_{\mathrm{batch}}G$ candidate rubrics \\
\bottomrule
\end{tabular}
\caption{RACE online model calls for one optimizer update. One rubric is shared by all $G$ policy responses in a prompt group. During a rubric update, each of the $G$ candidate rubrics is evaluated on both responses of the single cached ordered pair $(y^+,y^-)$ drawn from $\mathcal P$.}
\label{tab:race_step_compute}
\end{table*}

%% file: tab/hyperparameter.tex
\begin{table*}[t]
\centering
\small
\setlength{\tabcolsep}{0.5mm}
\begin{tabular}{lccccccccc}
\toprule
\textbf{Controller Setting}
& \textbf{GSM8K}
& \textbf{MBPP+}
& \textbf{BBH}
& \textbf{MMLU}
& \textbf{PopQA}
& \textbf{GPQA}
& \textbf{Zebra}
& \textbf{AGI-E}
& \textbf{Avg.} \\
\midrule

Default $(0.20,25,12)$
& \underline{95.9} & \textbf{68.7} & \textbf{72.3}
& \textbf{86.1} & \textbf{33.8} & \underline{56.1}
& \textbf{83.9} & \textbf{87.5} & \textbf{73.0} \\

Short Phase Targets $(0.20,10,5)$
& 95.9 & 67.6 & 66.4
& 85.2 & 32.7 & 54.1
& 82.0  & 87.1 & 71.4 \\

% Long Phase Targets $(0.20,50,40)$
% & & & & & & & & & \\

Lower Quarantine Target $(0.10,25,12)$
& 96.1
& 65.7
& 63.8
& 84.8
& 32.9
& 56.9
& 83.6
& 87.4
& 71.4 \\

\bottomrule
\end{tabular}
\caption{\textbf{Hyperparameter sensitivity.}
Each tuple denotes $(q^*, L^*, M^*)$, corresponding to the target quarantine
rate and the target policy- and rubric-phase lengths.}
\label{tab:hyperparameter_sensitivity}
\end{table*}